\documentclass[10pt, twocolumn, journal]{IEEEtran}

\usepackage{enumerate}
\usepackage{graphicx}
\usepackage{epsf}
\usepackage{verbatim}
\usepackage{amsmath}
\usepackage{amssymb}
\usepackage{amsfonts}
\usepackage{amsthm}
\usepackage{array}
\usepackage{makecell}
\usepackage{pifont}
\usepackage{algorithmic}
\usepackage[linesnumbered,ruled,vlined]{algorithm2e}
\theoremstyle{definition}

\usepackage{multirow} 
\usepackage{tabularx}
\usepackage{color}
\usepackage{url}
\usepackage{cite}
\usepackage{bm}

\DeclareMathOperator*{\argmax}{\arg\!\max}

\graphicspath{{Fig/},{fig/}}

\newcommand{\SYSFull}{skeleton-assisted learning-based clustering localization\xspace}
\newcommand{\SYS}{SALC\xspace}
\newcommand{\fig}{Fig.\xspace}
\newcommand{\RSSFullC}{RSS-Oriented Map-Assisted Clustering\xspace}
\newcommand{\RSSFull}{RSS-oriented map-assisted clustering\xspace}
\newcommand{\RSS}{ROMAC\xspace}
\newcommand{\REGFullC}{Cluster-based Online Database Establishment\xspace}
\newcommand{\REGFull}{cluster-based online database establishment\xspace}
\newcommand{\REG}{CODE\xspace}
\newcommand{\WKNN}{CsLE\xspace}
\newcommand{\WKNNFullC}{Cluster-scaled Location Estimation\xspace}
\newcommand{\WKNNFull}{cluster-scaled location estimation\xspace}

\newcommand{\NN}{CODE-NN\xspace}
\newcommand{\LR}{CODE-LR\xspace}

\usepackage[labelformat=simple]{subcaption}

\newcommand\numberthis{\addtocounter{equation}{1}\tag{\theequation}}

\begin{document}
\title{SALC: Skeleton-Assisted Learning-Based Clustering for Time-Varying Indoor Localization}


\author{\IEEEauthorblockN{An-Hung Hsiao, Li-Hsiang Shen$^\dagger$, Chen-Yi Chang, Chun-Jie Chiu, and Kai-Ten Feng}\\
\IEEEauthorblockA{Department of Electrical and Computer Engineering\\ National Yang Ming Chiao Tung University, Hsinchu, Taiwan\\
$^\dagger$California PATH, Institute of Transportation Studies, University of California at Berkeley, Berkeley, USA\\
Email: eric28200732.cm07g@nctu.edu.tw, shawngp3@berkeley.edu, zx57012tw@gmail.com, jack0502801.cm02g@mail.nctu.edu.tw, ktfeng@nycu.edu.tw
%\IEEEauthorrefmark{1}phtseng@mail.moj.gov.tw
}}

\maketitle

\begin{abstract}
Wireless indoor localization has attracted significant amount of attention in recent years. Using received signal strength (RSS) obtained from WiFi access points (APs) for establishing fingerprinting database is a widely utilized method in indoor localization. However, the time-variant problem for indoor positioning systems is not well-investigated in existing literature. Compared to conventional static fingerprinting, the dynamically-reconstructed database can adapt to a highly-changing environment, which achieves sustainability of localization accuracy. To deal with the time-varying issue, we propose a skeleton-assisted learning-based clustering localization (SALC) system, including \RSSFull (\RSS), \REGFull (\REG), and \WKNNFull (\WKNN). The SALC scheme jointly considers similarities from the skeleton-based shortest path (SSP) and the time-varying RSS measurements across the reference points (RPs). \RSS clusters RPs into different feature sets and therefore selects suitable monitor points (MPs) for enhancing location estimation. Moreover, the \REG algorithm aims for establishing adaptive fingerprint database to alleviate the time-varying problem. Finally, \WKNN is adopted to acquire the target position by leveraging the benefits of clustering information and estimated signal variations in order to rescale the weights from weighted k-nearest neighbors (WkNN) method. Both simulation and experimental results demonstrate that the proposed \SYS system can effectively reconstruct the fingerprint database with an enhanced location estimation accuracy, which outperforms the other existing schemes in the open literature.

%Using received signal strength (RSS) obtained from WiFi access points (APs) as the information of fingerprinting database is a widely utilized method in indoor localization. However, the time-variant problem for indoor positioning systems is not well-investigated in existing literature. Compared to conventional static fingerprinting, the dynamically-reconstructed database can adapt to a highly-changing environment which achieves sustainability of localization accuracy. To deal with the time-varying issue, we propose a skeleton-assisted learning-based clustering localization (SALC) system, including \RSSFull (\RSS), \REGFull (\REG), and \WKNNFull (\WKNN). The SALC scheme jointly considers similarities from the skeleton-based shortest path (SSP) and the time-varying RSS measurements across the reference points (RPs). Both simulation and experimental results demonstrate that the proposed \SYS system can effectively reconstruct the fingerprint database resulting in enhanced location estimation accuracy, which outperforms other existing schemes in open literature.

\end{abstract}


\section{Introduction}
%-------------------- Paper introduction -------------------------------

    For decades, the emerging location-based services (LBSs) have been promoted by telecom operators which significantly relies on acquiring the position of user equipment (UE) or target devices \cite{acm}. There exist abundant techniques to be adopted for LBS, such as global positioning system (GPS) \cite{GPS}, passive infrared (PIR) sensors \cite{PIR}, WiFi \cite{wifi,8468057}. LBS can be adopted in a variety of contexts including indoor/outdoor localization \cite{Hsiao-Chien} and human presence detection \cite{Frances}. Nowadays, as the life-oriented demands in public areas soar, indoor LBS capable of locating a particular person or monitoring the people flow has received considerable attention. However, GPS is not suitable for indoor LBS since their signals suffer from severe environmental degradation, including scattering and blockage, which leads to unpredictably low positioning accuracy. Therefore, short-range signal source such as WiFi becomes a potential candidate to be utilized in a complex indoor environment.

    WiFi-based localization system is widely applied in indoor positioning based on WiFi access points (APs) and portable devices using received signal strength (RSS) as information inference. The RSS is the signal information related to path-loss distance between the transmitter and receiver, which can be readily obtained from WiFi APs. Fingerprinting \cite{fingerprinting1,fingerprinting2,9744536,fingerprinting3,FL} is a widely-adopted positioning algorithm based on information of APs, which typically contains both offline measurement and online estimation phases. In the offline phase, the information is measured and collected at pre-defined locations so-called reference points (RPs) to establish database consisting of RSS from APs and geometric locations of RPs. During the online phase, real-time RSS will be received and matched to those from RPs in the offline-established database to estimate the target position with the aid of signal similarity features. In \cite{fingerprinting3}, the authors utilize complex channel information to overcome the problems, such as data loss, noise and interference in the fingerprint database and laborious offline training. The authors in \cite{FL} have proposed a solution for alleviating frequent data collection and improve privacy in two specifically defined scenarios. Hence, fingerprinting possesses lowered computational complexity and is capable of reflecting the multipath effects of non-line-of-sight in indoor environments \cite{survey}. Moreover, the weighted $k$-nearest neighbor (WkNN) algorithm \cite{PKNN} is employed to locate the desirable device during the online phase, which is calculated based on the $k$ largest weights among all RPs. Note that the weight implies the difference between real-time user's RSS and offline measured one in database. Consequently, it becomes important to investigate the factors disturbing RSS values which can result in acquiring faraway incorrect target location with similar RSS.

However, the RSS will also be affected by time-variation and human blockages, the authors in \cite{learning_based3} consider the relationship between RSS of monitor points (MPs) and of RPs collected in offline phase to construct an artificial neural network for position estimation. During the online phase, the real-time RSS values at RPs are predicted based on the collected data at MPs. In general, the intention of adopting MPs is to observe the environmental changes in specific areas by continuously collecting signal information. The MPs detect the variation of RSS so that it can provide immediate information to reconstruct a more precise real-time radio map. Moreover, the function of MP can be embedded into smartphones or mobile devices, which is considered relatively low-cost to monitor the RSS. Nevertheless, inappropriate deployed locations of MPs cause insufficient information for radio map establishment. Therefore, it is essential to design feasible schemes to cluster the RPs into several groups and to select the cluster head as MPs based on available signal sources and map information \cite{mapinfo}. The authors in \cite{affinity_propagation2} introduce the affinity propagation clustering algorithm which can deal with the RP clustering problem. The affinity propagation clustering process exchanges similarity by utilizing the responsibility and availability messages among the nodes in each iteration. Note that the responsibility message quantifies how well-suited the node serves as the exemplar to the other nodes; while availability message represents how appropriate the node selects the other node as its exemplar. After the clustering process, the nodes will be divided into several clusters and choose their own cluster head as the exemplar. The advantage of affinity propagation clustering is that the process only requires the similarity among data and is unnecessary to pre-define the number of clusters in most cases. The number of clusters can also be constrained by adjusting specific parameters. Hence, we can choose appropriate MPs from the RPs based on affinity propagation clustering by leveraging a designed similarity function with the aid of useful information.

   In this paper, to properly determine the locations of MPs, the proposed \RSSFull (\RSS) algorithm is designed based on affinity propagation clustering. With the aid of ML-enhanced techniques, \REGFull (\REG) is proposed to reconstruct the fingerprint database for solving the time-variation problem without recollecting RSS information of RPs. Moreover, we propose a \WKNNFull(\WKNN) algorithm to further improve the position estimation accuracy by computing adaptive weights based on the RSS's signal variance and cluster information.  Unlike conventional WkNN, \WKNN can localize the user accurately without selecting the farther RPs as candidates during the online phase.   The contributions of this paper are summarized as below.
    \begin{enumerate}
   \item We propose a \SYSFull (\SYS) system that optimizes RP clustering and MP selection in ROMAC, fingerprinting database reconstruction in CODE, and user positioning in CsLE in time-varying indoor environments. 
   
    \item \RSS utilizes affinity propagation to cluster RPs associated with the selected MPs as cluster heads. We jointly consider RSS differences, geometric relationship of RPs, and time-variation effect. Note that the clustering information from \RSS is utilized in both \REG and \WKNN for database reconstruction and accurate positioning.
    
   \item The proposed \REG algorithm aims for solving the time-variation issue by utilizing linear regression and neural network techniques to generate adaptive database for real-time radio maps. \WKNN aims for user positioning based on \RSS and \REG, which considers RSS variance from APs. It avoids noisy RSS signals and infeasible faraway RPs as candidate locations.
   
   \item Simulation results show that proposed \RSS appropriately clusters the RPs as well as chooses the MP cluster head. \REG can efficiently reconstruct a real-time radio map to solve time-variation. While, \WKNN reaches a higher localization estimation accuracy than existing methods. Moreover, real-time experiments are conducted to validate the effectiveness of our proposed \SYS system, with a higher positioning accuracy.
  \end{enumerate}

    The remaining of the paper is organized as follows. In Section \ref{Related}, we have investigated the related works. In Section \ref{section2}, the system architecture and flowchart of \SYS are demonstrated. The proposed \RSS, \REG, and \WKNN schemes are elaborated in Section \ref{section3}.  In Section \ref{section4}, the simulation and experimental results are discussed in detail. Conclusions are drawn in Section \ref{section5}.

\section{Related Work}\label{Related}
    Indoor localization has been an interesting research for decades, conventional RSS-based fingerprinting for indoor positioning is studied in \cite{RADAR,HORUS}. There exist critical bottlenecks restricting its large-scale implementations which consist of time-consuming and labor-intensive data collection process under severe wireless environmental influence, e.g., RSS suffers from dynamic environmental changes such as temperature, shadowing, and obstacles. In \cite{survey2} and \cite{survey3}, the authors have provided a comprehensive survey of existing fingerprinting methods and dynamic update techniques for radio maps. However, they did not consider the concept of MP deployment, which is capable of efficiently updating the radio map by selecting the appropriate MPs. As a result, online RSS measurement may significantly deviate from the fingerprint database established in the offline phase. However, the RSS will also be affected by time-variation and human blockage, which are not jointly considered in conventional schemes. Furthermore, static fingerprint database may be unreliable, which requires repeated data collection to maintain a satisfactory positioning accuracy. The works of \cite{CSE,path_loss1,path_loss2} have conceived different methods to solve the above-mentioned problems. For example, the authors in \cite{CSE} estimate the RSS at non-site-surveyed positions and utilize the support vector regression (SVR) to improve the resolution of the radio map. Existing researches calibrate the database mainly based on either distance-based path-loss models or interpolation method. However, the paper of \cite{loss_model1} has shown that it is difficult to adopt signal loss models in the complex and time-varying environments.

    Some existing works intend to modify the classical WkNN to improve localization accuracy by considering different factors \cite{WKNN2,RSS+CSI,WKNN3,RWKNN}. \cite{WKNN2} has proposed a feature-scaled WkNN, depending on the observation of different RSS values and distinguishability in geometrical distances. However, solely taking either RSS or variance \cite{RSS+CSI} into consideration is potentially insufficient to precisely estimate the user position in sophisticated indoor wireless environments. Therefore, the authors in \cite{WKNN3} have proposed a new weighted algorithm based on geometric distance of the RPs, and authors in \cite{RWKNN} also proposed a restricted WkNN by considering indoor moving constraints to reduce spatial ambiguity. However, the RSS will also be affected by time-variation and human blockage, which is not jointly considered in the conventional schemes. Consequently, it becomes important to investigate the factors disturbing RSS values, which can result in acquiring faraway incorrect target locations with similar RSS.

    Under such issues of complex indoor environments and the nonlinearity of radio map caused by time-varying effect, the state-of-the-art machine learning (ML) technique is capable of intelligently estimating user position and of dynamically establishing fingerprint database in an effective and efficient manner. The works in \cite{linear_regression2,linear_regression3} utilize different linear regression methods to calibrate the received signals in order to alleviate the variation of RSS. The works \cite{learning_based1,learning_based2,learning_based4} have proposed linear regression-enhanced methods to update the online radio map based on the observation of real-time received signals. In \cite{Transfer}, the authors have applied transfer learning to realize adaptive database construction with the aid of the arbitrarily deployed MPs. The authors in \cite{FL} have utilized the federated learning to address dynamic and heterogeneous data streams in indoor localization. In \cite{my2}, time-varying effect is taken into account with the aid of teacher-student learning. However, it requires laborious data collection as well as site-specific training. In \cite{FL, my2}, they both require a much more complex neural network and training mechanism as well as high-dimensional data, which may lead to time-consuming process. Moreover, the existing works did not jointly consider the time-variation and human blockage effects in the online phase. We propose the SALC to address the above-mentioned problem, which will be introduced in the following section in detail.

\section{System Architecture and Problem Formulation}\label{section2}

    \begin{figure}[t]
    \centering 
    \includegraphics[width=3.3in]{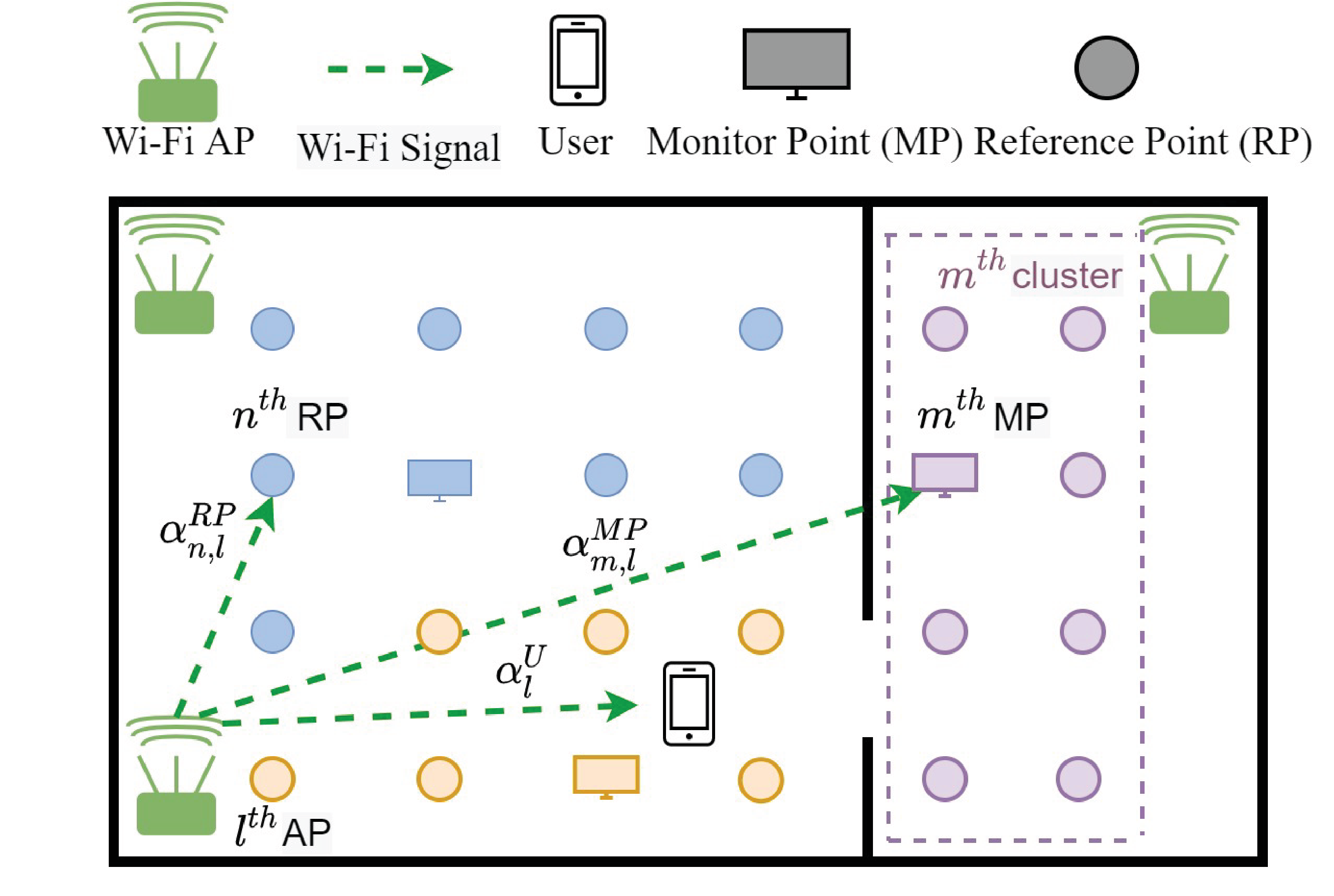}
    \captionsetup{font={footnotesize}}
    \caption{Network scenario for proposed \SYS system. It shows that $3$ APs are deployed in two rooms with three clusters labeled with different color, where the corresponding $3$ MPs are chosen as cluster heads to perform real-time monitoring of time-varying RSS signals.} \label{figure:Schematic}
    \end{figure}
    
    \fig\ref{figure:Schematic} illustrates the WiFi network scenario of indoor localization. In the proposed SALC system, we jointly consider RP clustering, MP deployment, adaptive fingerprinting database reconstruction, and user position prediction based on the RPs' time-varying RSS and geometrical information, which have not been considered in existing literature. The network is deployed with $N_{ap}$ APs, $N_{rp}$ RPs and $N_{mp}$ MPs, where $N_{mp}$ will be determined in the proposed SALC scheme in the next section. During the offline phase of fingerprinting, the measured RSS $\alpha_{n,l}^{RP}(t)$ is received from the $l^{th}$ AP on the $n^{th}$ RP at the $t^{th}$ time instant, where the location of the corresponding RP is also recorded. The offline database containing collected RSS measurements from all RPs is represented as 
    \begin{align}\label{eq:rss_measurement}
         \bm{\alpha}_{l}^{RP}(t) &= [ \alpha_{1,l}^{RP}(t), \dots , \alpha_{n,l}^{RP}(t), \dots , \alpha_{N_{rp},l}^{RP}(t) ]^\top, 
    \end{align}
    where $l \in [1,N_{ap}]$ and $n \in [1,N_{rp}]$ are the indexes of APs and RPs, respectively, and $\top$ is defined as transpose operation. Note that $\alpha_{n,l}^{RP}(t)$ is estimated based on the path-loss model specified in \cite{propagation_model}, which is expressed as 
    \begin{align}\label{eq:pathloss}
        \alpha_{n,l}^{RP}(t) = P_t - 20\log(f)+P_{d}\log(d_{n,l})-28,
\end{align}
where $P_t$ is the transmit power, $f$ is the operating frequency, $P_{d}$ is the path-loss coefficient depending on indoor environments, and $d_{n,l}$ is the distance between the $n^{th}$ RP and the $l^{th}$ AP. Note that $\alpha_{n,l}^{RP}(t)$ is in the unit of dB. The locations of MPs will be determined among RPs as cluster heads considering the similarity among RPs with respect to time, RSS, and geometric distance, which aim for monitoring the time-varying RSS at time $t$ in its own cluster. The measurement of MP's RSS can be given by 	
    \begin{align}\label{eq:mp_rss_measurement}
         \bm{\alpha}_{l}^{MP}(t) &= [ \alpha_{1,l}^{MP}(t), \dots , \alpha_{m,l}^{MP}(t), \dots , \alpha_{N_{mp},l}^{MP}(t) ]^\top, 
    \end{align}
where $m \in [1,N_{mp}]$ is the index of MPs and $\alpha_{m,l}^{MP}(t)$ is estimated based on the path-loss model the same as that in \eqref{eq:pathloss}. In the online phase, the adaptive fingerprinting database will be generated by applying both real-time received RSS from MPs and offline established RSS database. Therefore, the user's position can be estimated by matching its real-time RSS $\alpha^{U}_l(t)$ to that in the generated fingerprinting database. 
    
    \begin{figure}[t]
    \centering
    \includegraphics[width=3.3in]{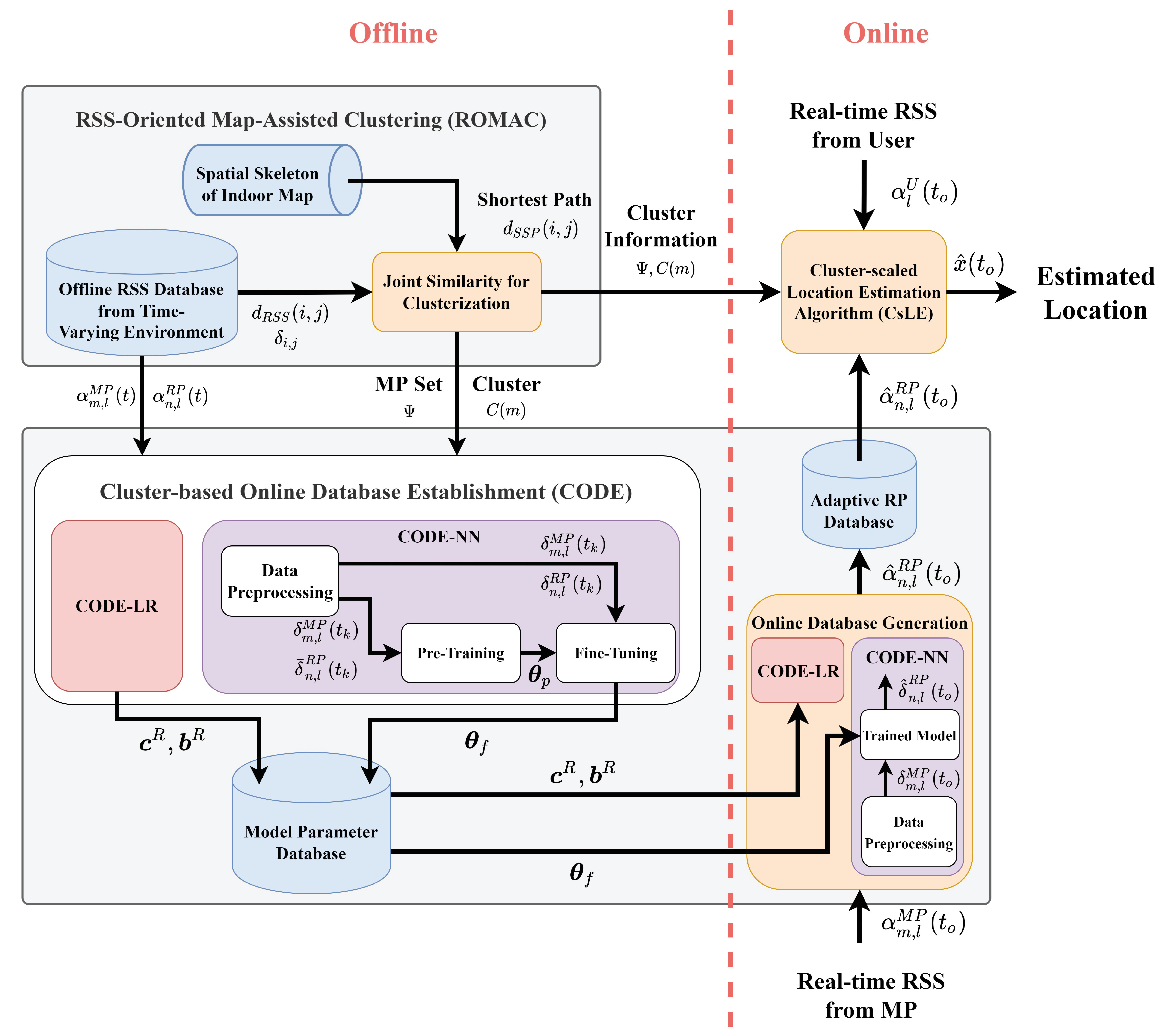}
    \captionsetup{font={footnotesize}}
    \caption{System flowchart of proposed \SYSFull (\SYS) which includes \RSS, \REG, and \WKNN dealing with RPs clustering, MPs selection, fingerprinting database reconstruction, and user positioning.} \label{figure:Flowchart}
    \end{figure}

    The flowchart of the proposed \SYS system is shown in \fig\ref{figure:Flowchart}, which is composed of three sub-algorithms, including \RSS, \REG, and \WKNN. The main target of \SYS is to deal with the time-variation issues in conventional fingerprinting database resulting in incorrect radio map matching, which consequently incurs a lower location estimation accuracy. By monitoring the environmental changes with the aid of MPs, the distribution of $\alpha^{RP}_{n,l}(t)$ can be well-estimated to establish the adaptive RP database as
    \begin{align}\label{eq:rp_problem}
        p\left(\alpha^{RP}_{n,l}(t_o)\right)={p\left(\alpha^{RP}_{n,l}(t)|\alpha^{MP}_{m,l}(t_o)\right)},
    \end{align}
where $t_o$ indicates the measurement time step in online phase. We need to obtain the RP's RSS value $\alpha^{RP}_{n,l}(t)$ that maximizes the distribution $p(\alpha^{RP}_{n,l}(t_o))$ on each RP. Therefore, based on (\ref{eq:rp_problem}), we can acquire more precise RSS on each RP which is formulated as
    \begin{align}\label{eq:MPM}
        \bm{\alpha}^{RP}_l(t_o) = M\left(\bm{\alpha}^{MP}_l(t_o)\right) =\argmax_{\alpha^{RP}_{n,l}(t)}p\left(\alpha^{RP}_{n,l}(t)|\alpha^{MP}_{m,l}(t_o)\right),
    \end{align}
    where $M \left( \bm{\alpha}^{MP}_l(t_o) \right): \mathbb{R}^{N_{mp}} \rightarrow \mathbb{R}^{N_{rp}}$ indicates the mapping function from MPs to RPs. We can observe from \eqref{eq:MPM} that it leads to a non-linear optimization problem due to the sophisticated indoor environment with ample multi-paths and blockages. The proposed \RSS in the upper-left of \fig\ref{figure:Flowchart} is adopted to select the MPs in \eqref{eq:MPM} to provide instant information of $\alpha^{MP}_{m,l}(t_o)$. Furthermore, the proposed \REG in the lower-left of \fig\ref{figure:Flowchart} is designed to generate the distribution of $\alpha^{RP}_{n,l}(t_o)$ by employing deep neural networks in order to solve the non-linear estimation problem in \eqref{eq:MPM}. We can then obtain the estimated user position $\hat{\bm{x}}(t_o)=[\hat{x}(t_o), \hat{y}(t_o)]^{\top}$ according to the received RSS $\bm{\alpha}^{U}(t_o)$ at its real position $\bm{x}=[x, y]^{\top}$, which is expressed as
    \begin{align}\label{eq:problem}
         \hat{\bm{x}}(t_o)=\argmax_{\forall\bm{x}}{p\left(\bm{x}|\bm{\alpha}^{U}(t_o)\right)},
    \end{align}
where $\bm{\alpha}^{U}(t_o)=[\alpha^{U}_1(t_o),\dots,\alpha^{U}_l(t_o),\dots,\alpha^{U}_{N_{ap}}(t_o)]$. The localization problem \eqref{eq:problem} is equivalent to the following formula as
    \begin{align}\label{eq:posterior}
         p\left(\bm{x}|\bm{\alpha}^{U}(t_o)\right) = \frac{p\left(\bm{\alpha}^{U}(t_o)|\bm{x}\right)p\left(\bm{x}\right)}{p\left(\bm{\alpha}^{U}(t_o)\right)},
    \end{align}
    where the posterior distribution $p(\bm{x}|\bm{\alpha}^{U}(t_o))$ can be derived based on Bayes' theorem \cite{Bayes}. The probability $p(\bm{x})$ follows the uniform distribution, which therefore can be ignored due to its proportionality property. Moreover, the distribution of RSS $\bm{\alpha}^{U}(t_o)$ is attainable because RSS is collected from the mobile device. Accordingly, maximizing likelihood $p\left(\bm{\alpha}^{U}(t_o)|\bm{x}\right)$ is equivalent to maximizing the radio map mapping function $R(\bm{x}):\ \mathbb{R}^{N_{rp}}\rightarrow\mathbb{R}^{2}$ which can be represented as
    \begin{align}\label{}
      R(\bm{x})=\argmax_{\bm{\alpha}^{U}(t)}{p\left(\bm{\alpha}^{U}(t)|\bm{x}\right)}\approx \sum_{n=1}^{N_{rp}}\alpha^{RP}_{n,l}(t)\delta(\bm{x}-\bm{x}_n),
    \end{align}
where $\delta(\cdot)$ is a delta function with value equal to $1$ if $\bm{x}=\bm{x}_n$, and $\bm{x}_n$ is the location of the $n^{th}$ RP. The radio map $R(\bm{x})$ is approximated from the summation of updated parameter $\alpha^{RP}_{n,l}(t)$ from \eqref{eq:rp_problem}.

\section{Proposed SALC System}\label{section3}

    The proposed \SYS system provides location estimation with the employment of fingerprinting and received RSS values from MPs. However, the data collection will suffer from the non-linear time-variation issue during both offline and online phases. To deal with this issue, it becomes crucial to design a non-linear MP-aided mapping function in order to reconstruct the real-time radio map. Therefore, we propose three sub-algorithms including \RSS, \REG, and \WKNN in the \SYS system to solve the above-mentioned issues, i.e., RPs clustering, MPs selection, fingerprinting database reconstruction, and user positioning. The \RSS is mainly designed to divide the RPs into clusters and select corresponding MPs, whereas \REG is adopted to generate the distribution of $\bm{\alpha}^{RP}_{l}(t)$ in order to solve the non-linear problem in \eqref{eq:MPM}. The proposed \WKNN scheme aims for estimating user's position according to the adaptive database constructed by \REG and the clustering information obtained from \RSS.
    
    \subsection{\RSSFullC (\RSS)}
The proposed ROMAC scheme is designed based on an unsupervised learning approach, which can classify the unlabeled data into different groups based on attainable RSS features. As mentioned in Section \ref{Related}, the existing methods for the MP deployment have not jointly considered the signal strength, map information and time-varying effect. In this subsection, we will introduce how \RSS jointly considers all important factors, including RSS measurements, skeleton-based shortest path (SSP), and time-variation characteristics to conduct the clustering process. Furthermore, deployment of MPs is also determined by selecting the cluster head. \RSS is designed based on the affinity propagation \cite{affinity_propagation1}, which only requires the similarity feature among RPs, without the need to pre-define the number of clusters. The self-defined similarity consists of three key factors including the amplitude difference of RSS among RPs, layout of RPs, and time-variance of RSS. The amplitude difference of RSS represents potential signal fading and blockages. The indoor layout provides the position knowledge for the SSP, whilst the time-variance of RSS reflects the time-varying effect of signals. Based on the observed received signal $\bm{\alpha}_{l}^{RP}(t)$, the difference of RSS between the $i^{th}$ and $j^{th}$ RPs is defined as
    \begin{align}\label{drss}
    d_{RSS}(i,j) = \frac{1}{N}\sum_{k=1}^{N} \left|\alpha_{i,l}^{RP}(t_k)-\alpha_{j,l}^{RP}(t_k) \right|,
    \end{align}    
where $N$ is the considered time interval, $t_k$ is the time index for $k \in [1, N]$ and the notion of $|x|$ represents the absolute value of $x$. Notice that the difference between the $i^{th}$ and $j^{th}$ RP's RSS is related to its Euclidean distance, where a smaller value of $d_{RSS}(i,j)$ represents a higher similarity level between RPs. 

    \begin{figure}[t]
    \centering
    \includegraphics[width=0.4\textwidth]{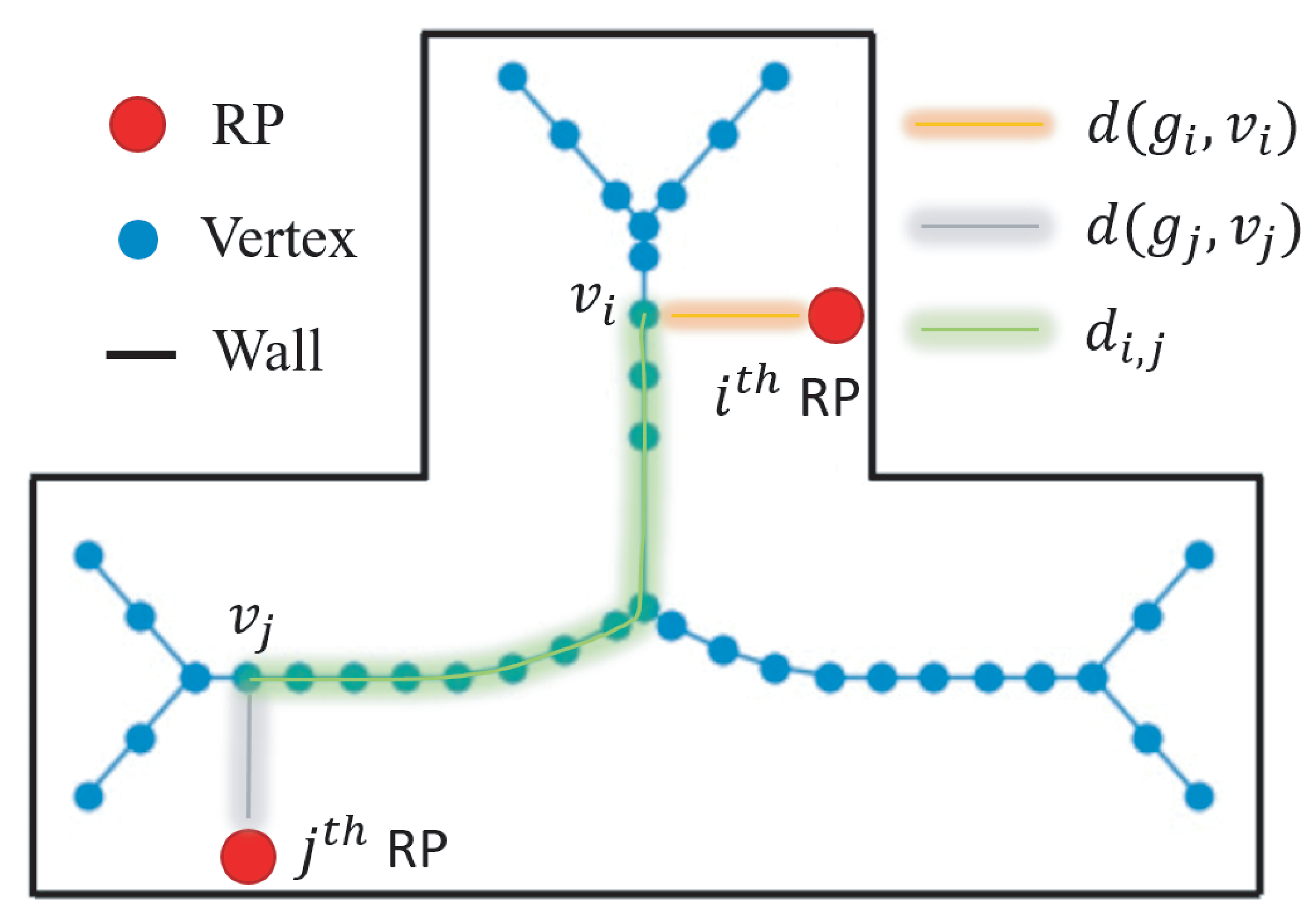}
    \captionsetup{font={footnotesize}}
    \caption{The schematic diagram of spatial skeleton and RPs. The white area is walkable area and the black lines are the walls. The red points in walkable area are RPs, the blue lines represent the skeleton and the blue points are the vertices of skeleton. The shortest path $d_{SSP}(i,j)$ between the $i^{th}$ and the $j^{th}$ RPs is also illustrated.} \label{figure:skeleton}
    \end{figure} 
    
    \fig\ref{figure:skeleton} shows an exemplified layout of an indoor environment, where the white area is walkable area and the black lines are the walls. The red points are RPs to be estimated and blue lines and vertices are formed by the SSP scheme adopted from \cite{Skeleton_skeleton} to acquire the spatial skeleton, which provides a compact map of the shortest paths among RPs. The vertices of SSP-skeletons are calculated based on generalized Voronoi diagram (GVD) technique \cite{GVD} as an enhancement of conventional Voronoi diagram in order to partition a plane into several cell-like regions. The output of GVD contains edges and vertices as demonstrated in blue lines and points in \fig\ref{figure:skeleton}, respectively. We define $\bm{D}$ as an SSP matrix which is given by $\bm{D} = [d_{v,w}],\quad \forall v,w\in [1,N_v]$, where $d_{v,w}$ is the shortest path from the $v^{th}$ to $w^{th}$ vertices, and $N_v$ is the total number of vertices. Therefore, as shown in \fig\ref{figure:skeleton}, the spatial distance of RPs can be derived from the shortest path between the $i^{th}$ and the $j^{th}$ RPs as
    \begin{align}\label{dssp}
    d_{SSP}(i,j) = d(g_{i},v_{i}) + d_{i,j} + d(g_{j},v_{j}), 
    \end{align}
where $i,j \in [1,N_{rp}]$ are the indexes of RPs, $v_{i}$ and $v_{j}$ are respectively the nearest vertex of the $i^{th}$ and the $j^{th}$ RP. $g_{i}$ and $g_{j}$ are the positions of $i^{th}$ and $j^{th}$ RPs, respectively. $d(g_{i},v_{i})$ denotes the Euclidean distance between $i^{th}$ RP and vertex $v_{i}$, whilst $d_{i,j}$ denotes the  shortest path between the $i^{th}$ and $j^{th}$ vertices. Therefore, $d_{SSP}(i,j)$ reflects the spatial relationship between the $i^{th}$ and $j^{th}$ RPs in an SSP-aided layout. Note that smaller value of $d_{SSP}(i,j)$ indicates higher spatial relationship.
    
    Moreover, the long-term difference of RSS time-varying between empty and crowded areas is considered. The difference between the $i^{th}$ and the $j^{th}$ RP is derived from
    \begin{equation}\label{dtimevarying}
    \begin{aligned} 
    \delta_{i,j}\!=\!\left|\sum_{l=1}^{N_{ap}}\!\sum_{k=1}^{N}[\alpha_{i,l}^{RP}(t_k)\!-\!\alpha_{i,l}^{RP}(t_e)]\!-\![\alpha_{j,l}^{RP}(t_k)\!-\!\alpha_{j,l}^{RP}(t_e)]\right|^{-1},
    \end{aligned} 
    \end{equation}
    where $\alpha_{i,l}(t_e)$ is RSS measured on the $i^{th}$ RP from $l$ detectable APs which is considered as a reference RSS obtained at the time instant $t_e$, e.g., RSS acquired from an empty area. The parameter $\delta_{i,j}$ reflects the distinct difference of time-varying effect between the $i^{th}$ and $j^{th}$ RPs. Notice that a higher value of $\delta_{i,j}$ indicates that the RPs are less susceptible to time-variation effect. For example, consider the case that $\delta_{1,2} = 1/(11-1)=0.1$ with $[\alpha_{1,l}^{RP}(t_k)-\alpha_{1,l}^{RP}(t_e)]=11$ and $[\alpha_{2,l}^{RP}(t_k)-\alpha_{2,l}^{RP}(t_e)]=1$; while $\delta_{3,4} = 1/(2-1)=1$ with $[\alpha_{3,l}^{RP}(t_k)-\alpha_{3,l}^{RP}(t_e)]=2$ and $[\alpha_{4,l}^{RP}(t_k)-\alpha_{4,l}^{RP}(t_e)]=1$. It can be intuitively observed that the area around RPs $1$ and $2$ with $\delta_{1,2} =0.1$ is more susceptible to time-varying effect, e.g., a crowded room, comparing to that for RPs $3$ and $4$ with $\delta_{3,4} = 1$, e.g., an empty room. The RPs that are highly influenced by time-variation, i.e., smaller $\delta_i,j$ will be treated with larger similarity, since the proposed MPs are designed to resist time-varying effect.
     
    According to RSS, SSP and the time-varying effect in \eqref{drss}, \eqref{dssp} and \eqref{dtimevarying}, respectively, the joint similarity $s_{i,j}$ between the $i^{th}$ and the $j^{th}$ RPs can be formulated, which is also shown at top-left of \fig\ref{figure:Flowchart}, as
    \begin{equation}
    \begin{aligned}\label{similarity}
    s_{i,j} = - \bm{\omega}\cdot[d_{RSS}(i,j),d_{SSP}(i,j),\delta_{i,j}]^\top,
    \end{aligned}
    \end{equation}
    where $\bm{\omega} = [\omega_{RSS}, \omega_{SSP},\omega_{\delta}]$ represents the important weights of RSS, SSP and time-varying effect. Notice that we impose the negative sign on the factors in $\eqref{similarity}$ to reflect smaller values of those three factors resulting in larger joint similarity. We denote $\bm{S}_j$ as the joint similarity between the $j^{th}$ RP and the others, which is defined as
    \begin{align}
        \bm{S}_j = [s_{1,j},\dots s_{i,j}, \dots, s_{N_{rp},j}],
    \end{align}
where $ \forall i \ne j$, and $i,j \in [1, N_rp]$. Furthermore, we define the preference to represent the self-similarity of the $j^{th}$ RP as
    \begin{align}\label{self-similarity}
    s_{j,j} = M_d(\bm{S}_j), 
    \end{align}  
where $M_d(\cdot)$ is the median function averaging all elements of a matrix. The preference indicates the probability of a specific RP becoming a cluster exemplar, i.e., it is selected as the corresponding MP. Accordingly, a lower preference value of an RP indicates that it behaves similar to the other RPs, which means that it possesses a lower chance to be selected as a cluster exemplar.
    
    Based on the SSP-aided map and similarity definition, the proposed \RSS considers two types of messages among RPs including responsibility message $r_{i,j}$ and availability message $a_{i,j} \,\forall \ i,j \in [1,N_{rp}]$. The message will be exchanged iteratively in order to derive the prioritized responsibility and availability messages. The responsibility $r_{i,j}$ is sent from the $i^{th}$ RP to the $j^{th}$ RP, which is defined as
    \begin{align}\label{responsibility}
        r_{i,j} &= s_{i,j} - \max_{j' \neq j}\{a_{i,j'}+s_{i,j'}\},
    \end{align}
    where $\max \{ \cdot \}$ function gives the maximum value among input elements. The availability $a_{i,j}$ is then sent from the $j^{th}$ RP to the $i^{th}$ RP represented by
    \begin{align}\label{availability}
    a_{i,j} &= 
    \begin{cases}\min \{0,r_{j,j}+\sum_{i' \neq i,j}\max \{0,r_{i',j}\}\},\quad  & 
      \forall i\neq j,\\ 
      \sum_{i' \neq j}\max \{0,r_{i',j}\}, \quad  & \forall i=j,
    \end{cases} 
    \end{align} 
    where $\max \{ \cdot,\cdot \}$ will choose a larger element as the output and $\min  \{ \cdot,\cdot \}$ will choose a smaller one. $\eqref{responsibility}$ and $\eqref{availability}$ show that higher $r_{i,j}$ representing that the $j^{th}$ RP is more appropriate to serve as the candidate exemplar for the $i^{th}$ RP, whereas higher $a_{i,j}$ indicates that the $i^{th}$ RP has higher tendency to select the $j^{th}$ RP as its exemplar. The exemplar will be determined after both responsibility $r_{i,j}$ and availability $a_{i,j}$ matrix are updated. Consequently, the $j^{th}$ RP with the maximum value of $(r_{i,j}+a_{i,j})$ will be selected as the cluster exemplar MP. The set of exemplars $\Psi$ can be derived as
    \begin{align}{
        \Psi = \{\mu | \mu = \argmax_{\forall i,j \in [1,N_{rp}]} \{r_{i,j}+a_{i,j}\}\}.
        \label{MP}
        } 
    \end{align} 
Notice that the number of MPs can therefore be determined as $N_{mp} = \text{rank}(\Psi)$ according to the proposed ROMAC scheme. Furthermore, for the $i^{th}$ RP, its exemplar MP is selected from the exemplar set $\Psi$ as
    \begin{align}
        E(i)= \argmax_{j \in \Psi} \{r_{i,j}+a_{i,j}\}.
    \end{align} 
Alternatively, the $m^{th}$ RP as the $i^{th}$ RPs' exemplar MP is defined as the set of
    \begin{align}{
        C(m)= \{i|E(i)=\mu\}, \label{cluster}
        }
    \end{align} 
where $i \in [1,N_{rp}]$ is the index of RP in the $m^{th}$ cluster. Note that the parameter $\mu$ denotes the selected exemplar MP's actual RP's index number, whilst $m$ is defined as the re-ordered index of $\mu$ for the ease of representation in the following design.  For example, consider the case that $\Psi=\{\mu | \mu = 6,14,22,34\}$, the first cluster's exemplar MP with $m=1$ becomes $C(1)=\{i|E(i)=6\}$; while the second MP with $m=2$ is $C(2)=\{i|E(i)=14\}$. The iterations will be executed until the cluster set $C(m)$ becomes unchanged. As illustrated in \fig\ref{figure:Flowchart}, both the MP set $\Psi$ in (\ref{MP}) and the cluster set $C(m)$ in (\ref{cluster}) will be utilized as the inputs of the following \REG scheme. The concrete procedure of \RSS is provided in Algorithm \ref{alg1}.

	\begin{algorithm}[t]
	 \SetKwInOut{Input}{Input}
	 \SetKwInOut{Output}{Output}
	 \Input{Similarity $s_{i,j} = - \bm{\omega}\cdot[d_{RSS}(i,j),d_{SSP}(i,j),\delta_{i,j}]^{\top}$}
	 \Output{Exemplar set $\Psi$, cluster exemplar $E(i)$, cluster set $C(m)$}
	 Initialize responsibility $r_{i,j}$ and availability $a_{i,j}$ to 0\\
	 \While{the cluster set $C(m)$ is changed}{
	  Update responsibilities: $r_{i,j} \leftarrow s_{i,j} - \max_{j' \neq j}\{a_{i,j'}+s_{i,j'}\}$\\
	  Update availabilities: $a_{i,j} \leftarrow  \begin{cases}\min \{0,r_{j,j}+\sum_{i' \neq i,j}\max \{0,r_{i',j}\}\}\quad \ \ & 
	      \forall i\neq j,\\ 
	      \sum_{i' \neq j}\max \{0,r_{i',j}\} \quad \ \ & \forall i=j,
	    \end{cases} $\\
	  Update exemplar set: $\Psi \leftarrow \{\mu | \mu = \argmax_{\forall i,j \in [1,N_{rp}]} \{r_{i,j}+a_{i,j}\}\}$\\ 
	  Select MP for $i^{th}$ RP from exemplar set: $E(i) \leftarrow \argmax_{j \in \Psi} \{r_{i,j}+a_{i,j}\}$\\
	  Update cluster set: $C(m) \leftarrow \{i|E(i)=\mu\}$
	 }
	 \caption{\RSSFull (\RSS)}\label{alg1}
	\end{algorithm}

    \subsection{\REGFullC (\REG)}

In the proposed CODE scheme, both the linear regression (CODE-LR) and neural network (CODE-NN) schemes are designed in \fig\ref{figure:Flowchart}. CODE has addressed expired fingerprinting database issue caused by the time-varying effects. The radio map information can be predicted by adopting either regression or neural network models acting as the database generator. The proposed \LR scheme acquires the distribution in \eqref{eq:rp_problem} considering the linearity property of signal strength, i.e., RSS is approximately inversely proportional to the distance between the transmitter and receiver in a linear manner. To further account for nonlinear effects caused by indoor signal fading or moving objects, the conceived \NN scheme improves the accuracy of database prediction by extracting latent information from a deep neural network model.

    \subsubsection{\LR} 

    The proposed \LR algorithm is designed based on SVR \cite{SVR} to conduct online database construction. We intend to develop a predictive modeling technique investigating the relationship between dependent and independent variables to represent the measured RSS on MPs and RPs, respectively. The regression models of RPs will be trained considering the long-term RSS information in the offline database construction phase. During the online phase, the coefficients of regression models are extracted to establish the fingerprint database for online matching process. Note that the overhead of time-consuming offline database construction can therefore be reduced with the assistance of online adjustment to maintain satisfactory positioning accuracy of \SYS.
    
    Under time-varying environment, the RSS of each RP varies individually when environment changes. However, with the aid of cluster information $C(m)$ and deployed MPs $\Psi$ obtained from \RSS, we are capable of observing real-time RSS value. We notice that RPs in certain cluster behave similarly to their corresponding cluster exemplar MP, where the similarity pattern can be acquired via proposed \LR. Considering that RP $n \in C(m)$ and MP $m \in \Psi$ represent the cluster head of RPs $n \in C(m)$, the estimated RSS $\hat{\alpha}^{RP}_{n,l}(t_k)$ from the $l^{th}$ AP can be calculated by  
    \begin{align}
    \hat{\alpha}_{n,l}^{RP}(t_k)  = c_{n,l}^R\alpha_{m,l}^{MP}(t_k)  + b_{n,l}^R,
    \end{align}
    where $c_{n,l}^R$ and $ b_{n,l}^R$ are coefficient and bias, respectively, of regression in \LR. The loss function is modeled as
    \begin{align}
    J_{n,l} = \frac{1}{2}\sum_{k=1}^{N}\left(\hat{\alpha}_{n,l}^{RP}(t_k) - \alpha_{n,l}^{RP}(t_k)\right)^2.
    \end{align}
To iteratively update the weights of regression, the stochastic gradient descent is adopted as
    \begin{align}
    b_{n,l}^R &= b_{n,l}^R - \eta\frac{\partial J_{n,l}}{\partial b_{n,l}^R}
    = b_{n,l}^R - \eta\left[\hat{\alpha}_{n,l}^{RP}(t_k) - \alpha_{n,l}^{RP}(t_k)\right],\\
    c_{n,l}^R & = c_{n,l}^R - \eta\frac{\partial J_{n,l}}{\partial c_{n,l}^R}\nonumber\\
    & = c_{n,l}^R - \eta\left[\alpha_{m,l}^{MP}(t_k)\cdot\left(\hat{\alpha}^{RP}_{n,l}(t_k) - \alpha_{n,l}^{RP}(t_k)\right)\right],
    \end{align}
    where $\eta$ is the learning rate. The iteration will execute until the coefficient remains unchanged. Notice that we build a regression model for every RP, and the complete coefficient and bias sets can be represented as $\bm{c}^R=\{c_{n,l}^R \mid \forall n \in [1,N_{rp}],\forall l \in [1,N_{ap}] \}$ and $\bm{b}^R=\{ b_{n,l}^R \mid \forall n \in [1,N_{rp}],\forall l \in [1,N_{ap}] \}$. These two parameter sets will be obtained from the model training process of CODE-LR, as shown in the bottom-left part of \fig\ref{figure:Flowchart} and saved in the offline model parameter database.
    
    Furthermore, during the online phase, the online database generation will compute real-time RSS for RPs $\hat{\alpha}_{n,l}^{RP}(t_o)$ as shown in the bottom-right part of \fig\ref{figure:Flowchart} as
    \begin{align}
    \hat{\alpha}_{n,l}^{RP}(t_o)  = c_{n,l}^R\alpha_{m,l}^{MP}(t_o)  + b_{n,l}^R,
    \end{align}
    where $t_o$ denotes the time instant at online phase. $c_{n,l}^R$ and $ b_{n,l}^R$ are respectively acquired from $\bm{c}^R$ and $\bm{b}^R$ in model parameter database at the offline stage. Consequently, the RSS values from all RPs $\hat{\alpha}_{n,l}^{RP}(t_o)$ are computed and stored in the adaptive RP database in order to reconstruct the real-time radio map, which will be utilized for location estimation.
    
    \subsubsection{\NN}
    The linear relationship between the RSS values of MPs and RPs considered in \LR may be impractical in realistic system due to complicated indoor wireless environments. Therefore, \NN scheme is proposed to perform nonlinear training and mapping. The training process is divided into pre-training and fine-tuning phases. In the pre-training phase, the loss function and back-propagation will be computed by a cluster-averaged target in order to take cluster information into account to represent various similarity levels between RPs. In the fine-tuning phase, the pre-trained parameters will be loaded as initial parameters associating with the unaveraged target function in order to guarantee feasible prediction results.

    \begin{figure}[t]
    \centering
    \includegraphics[width=0.47\textwidth]{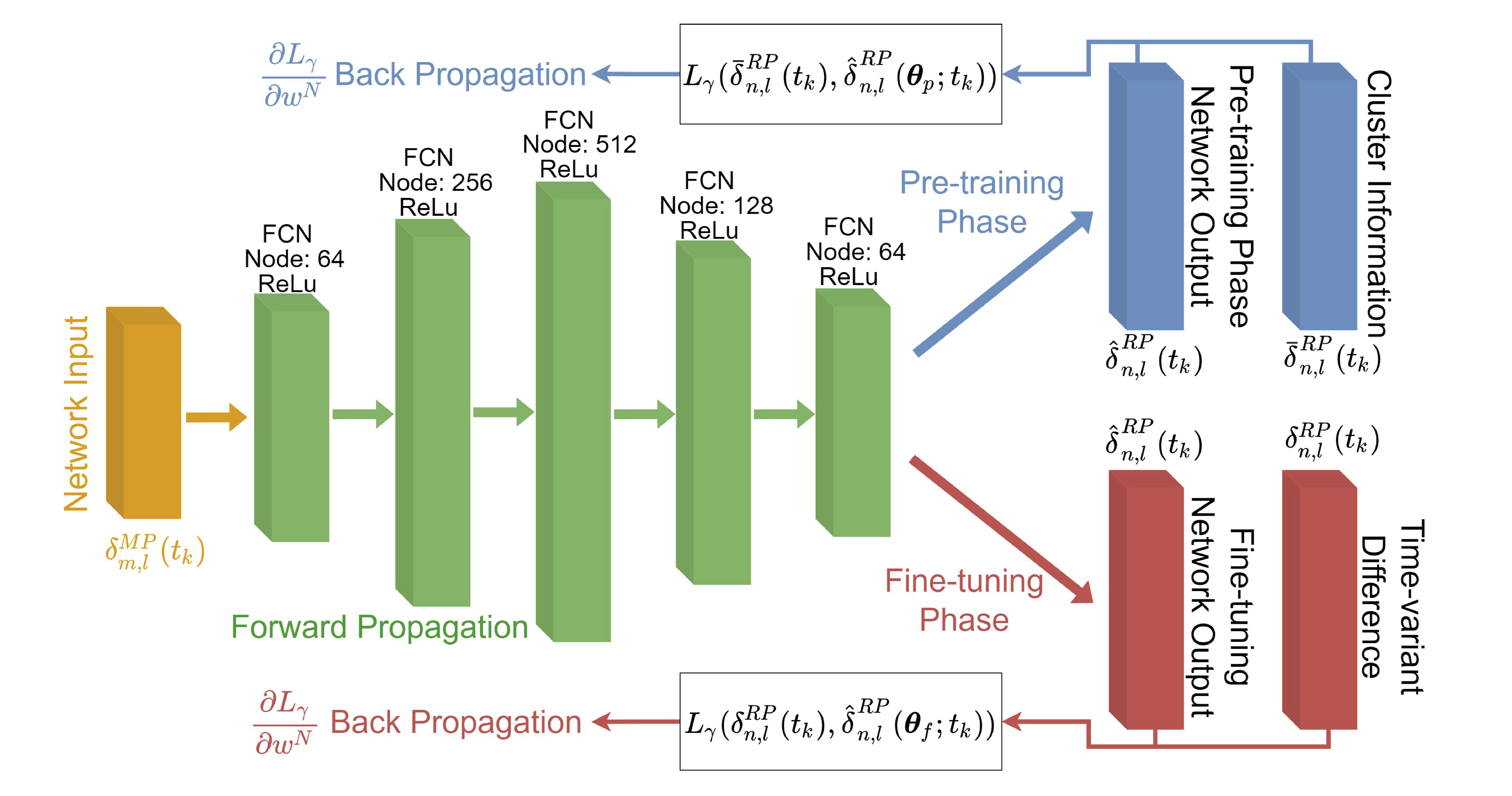}
    \captionsetup{font={footnotesize}}
    \caption{Network model for proposed \NN scheme.} \label{figure:NN}
    \end{figure} 
    
     First of all, data preprocessing as shown in \NN block of \fig\ref{figure:Flowchart} is applied in order to consider the time-varying effects. With the same cluster's RP $n \in C(m)$ and MP $m \in \Psi$, we subtract RSS from the $l^{th}$ AP in the initial environment $\alpha_{n,l}^{RP}(t_e)$ from that at time $t_k$, i.e., $\alpha_{n,l}^{RP}(t_k)$, to obtain the time-variant difference as
    \begin{align}
        \delta_{n,l}^{RP}(t_k) = \alpha_{n,l}^{RP}(t_k) - \alpha_{n,l}^{RP}(t_e). %\label{deltaRP}
    \end{align} 
Similarly, the difference of time-varying RSS obtained at the $m^{th}$ MP $\delta^{MP}_{m,l}(t_k)$ can be defined as
    \begin{align}
        \delta_{m,l}^{MP}(t_k) = \alpha_{m,l}^{MP}(t_k) - \alpha_{m,l}^{MP}(t_e).
    \end{align}

\fig\ref{figure:NN} shows the network architecture of \NN including the pre-training and fine-tuning phases. The network is constructed by the input and output layers with dimensions $N_{mp}$ and $N_{rp}$, respectively. Five fully connected neural network (FCN) layers are chosen as the hidden layers with the corresponding number of network nodes. Notice that the RSS difference $\delta^{MP}_{m,l}(t_k)$ acquired from all $N_{mp}$ MPs will be served as the inputs to FCN to provide the learning mechanism with an integrated fashion. The model output $\hat{\delta}^{RP}_{n,l}(t_k)$ can be computed by the forward propagation from hidden layers shown as the green part in \fig\ref{figure:NN}, which is represented as
    \begin{align} \label{deltaRP}
    \hat{\delta}^{RP}_{n,l}=f^{(5)}\left(\sum^{D_{5}}_{m=1}\dots f^{(1)}\left(\sum^{D_1}_{m=1}w^{(1)}_{n,m}\delta^{MP}_{m,l}+b^{(1)}_{n}\right)+b^{(5)}_{n}\right),
    \end{align}      
where $f^{(h)}(\cdot)$ is the activation function and $D_h$ is the number of nodes of the $h^{th}$ hidden layer for $h=1$ to $h=5$. The dimension of hidden layer's weight $w^{(h)}_{n,m}$ is $D_{h}\times D_{h-1}$ for $h=2$ to $5$ and $D_{1}\times N_{mp}$ for $h=1$, whilst that for the bias $b^{(h)}_n$ is $D_{h} \times1$ for all $h$.

    In the pre-training phase, the goal is to provide initial parameter of every node, which can help the model find the solution rapidly, i.e., to reduce the iterations of training process. After the data preprocessing in \fig\ref{figure:Flowchart}, the RSS difference $\delta^{RP}_{n,l}(t_k)$ at the $k^{th}$ time instant from (\ref{deltaRP}) are averaged within each cluster since the RPs in the same cluster have similar trend in time-variation. With the aid of cluster information $\Psi$ and $C(m)$, the averaged target can be represented as
\begin{align}
        \bar{\delta}^{RP}_{n,l}(t_k) = \frac{1}{N_m}\sum_{n=1}^{N_m}{\delta_{n,l}^{RP}(t_k)}, \label{deltaPrime}
\end{align}
where $N_m$ is the number of RPs in the $m^{th}$ cluster and $n$ is the index of RP, $\forall n \in C(m)$. Compared to utilizing $\delta_{n,l}^{RP}(t_k)$, choosing $\bar{\delta}^{RP}_{m,l}(t_k)$ in (\ref{deltaPrime}) as the target in pre-training phase can reduce computational complexity of loss function. Therefore, the loss function is designed between the target $\bar{\delta}^{RP}_{n,l}(t_k)$ and the output $\hat{\delta}^{RP}_{n,l}(\bm{\theta}_p^{\tau};t_k)$ as
    \begin{align*}\label{loss1}
        &L_\gamma\left(\bar{\delta}^{RP}_{n,l}(t_k),\hat{\delta}^{RP}_{n,l}(\bm{\theta}_p^{\tau};t_k)\right)\\
        &=\left\{
        \begin{footnotesize}
        \begin{aligned}
             &\frac{1}{2}\!\left[\bar{\delta}^{RP}_{n,l}(t_k)\!-\!\hat{\delta}^{RP}_{n,l}(\bm{\theta}_p^{\tau};t_k)\right]^2, \mbox{ for}
            \left|\bar{\delta}^{RP}_{n,l}(t_k)\!-\!\hat{\delta}^{RP}_{n,l}(\bm{\theta}_p^{\tau};t_k) \right| \leq \gamma,  \\
             &\gamma\left|\bar{\delta}^{RP}_{n,l}(t_k)-\hat{\delta}^{RP}_{n,l}(\bm{\theta}_p^{\tau};t_k)-\!\frac{1}{2}\gamma^2 \right |, \mbox{ otherwise},
            \end{aligned}
        \end{footnotesize}
        \right.
    \numberthis
    \end{align*}
where $\hat{\delta}^{RP}_{n,l}(\bm{\theta}_p^{\tau};t_k)$ denotes the model output predicted by the parameter at the $\tau^{th}$ training iteration during the pre-training phase. We select Huber loss \cite{huber} as the loss function in order to eliminate the effects of outliers by setting the threshold $\gamma$. The gradient descent method is utilized to search the optimum of the loss function during back propagation as shown in \fig\ref{figure:NN}. The gradient can be updated iteratively as 
    \begin{align}{\label{pretraining}}
        \bm{\theta}_p^{\tau+1}=\bm{\theta}_p^{\tau}-\eta\cdot\nabla L_\gamma\left(\bar{\delta}^{RP}_{n,l}(t_k),\hat{\delta}^{RP}_{n,l}(\bm{\theta}_p^{\tau};t_k)\right),
    \end{align}
where $\bm{\theta}_p^{\tau}$ is the parameter including the weight and bias during the pre-training phase, $\eta$ is learning rate, and $\nabla L_\gamma(\cdot,\cdot)$ is the first-order derivative of the loss function.
    
    After the pre-training phase, the following fine-tuning phase will be performed as shown in both Figs. \ref{figure:Flowchart} and \ref{figure:NN}. The parameter $\bm{\theta}_p$ will be saved and treated as the initial parameter for fine-tuning process, i.e., $\bm{\theta}_f^0 = \bm{\theta}_p^T$, where $\bm{\theta}_f^0$ and $\bm{\theta}_p^T$ represent the initial network parameter during fine-tuning phase and the network parameter updated at the last iteration $T$ in pre-training phase, respectively. Notice that the architecture of neural network and the number of nodes in fine-tuning phase are designed to be the same as those in the pre-training phase, where the loss function in fine-tuning can be acquired by replacing the averaged target $\bar{\delta}^{RP}_{n,l}(t_k)$ in (\ref{loss1}) with $\delta_{n,l}^{RP}(t_k)$ for each RP. Note that we also update the weights and bias of neural networks via the gradient descent method as that in \eqref{pretraining} by substitute $\bm{\theta}_p^{\tau}$ with $\bm{\theta}_f^{\tau}$. With the initial parameter acquired from the pre-training phase, the gradient descent can converge faster to find the optimum solution. 
    
    After completion of  pre-training and fine-tuning phases, the parameter $\bm{\theta}_f$ will be saved into the model parameter database to reconstruct the real-time radio map $\hat{\alpha}^{RP}_{n,l}(t_o)$ at the online phase. As shown in the lower-right of \fig\ref{figure:Flowchart}, the time-variation $\hat{\delta}^{RP}_{n,l}(t_o)$ is predicted based on the online monitored MP's RSS difference $\delta^{MP}_{m,l}(t_o)$ and the trained model parameter $\bm{\theta}_f$ via (\ref{deltaRP}). Once the network completes the calculation in online phase and outputs the prediction of $\hat{\delta}^{RP}_{n,l}(t_o)$, the element stored in the adaptive RP database in the online phase can be represented as
    \begin{align}
        \hat{\alpha}_{n,l}^{RP}(t_o) =  \hat{\delta}^{RP}_{n,l}(t_o)+\alpha_{n,l}^{RP}(t_e).
    \end{align} 
Note that $\hat{\alpha}_{n,l}^{RP}(t_o)$ represents the real-time adaptive radio map, which will be used to estimate the user position by the following \WKNN algorithm. Different from the linear property of \LR, \NN can solve the nonlinear problem of radio signal propagation such as human signal-blocking based on model training and adaptation.
  
    \subsection{\WKNNFullC (\WKNN)} 
    In order to accurately estimate the user position in the online phase, the proposed \WKNN scheme takes into account the signal variance caused by time-varying effects. As shown in the top-right part of \fig\ref{figure:Flowchart}, \WKNN is implemented based on real-time RSS collected from user $\alpha_{l}^{U}(t_o)$, cluster information with RP $n \in C(m)$ acquired from \RSS, and the adaptive RSS value $\hat{\alpha}_{n,l}^{RP}(t_o)$ for the $n^{th}$ RP via \REG. The modified Euclidean distance (MED) of online received RSS values from the $l^{th}$ AP between the $n^{th}$ RP and user is derived as
    \begin{align}
    \label{distance}
        d_{n,l}(t_o) = \sum_{m=0}^{M}\frac{\hat{\alpha}_{n,l}^{RP}(t_o)-\alpha_{l}^{U}(t_o)}{\sigma_{m}\left(\hat{\alpha}_{n,l}^{RP}(t_o)\right)},
    \end{align}
    where $\sigma_{m}(\cdot)$ is the weight scaling function that gives real-time estimated standard deviation of RSS from the $n^{th}$ RP within $m^{th}$ cluster. Higher value of $\sigma_{m}(\hat{\alpha}_{n,l}^{RP}(t_o))$ indicates that the $n^{th}$ RP's RSS from the $l^{th}$ AP is less reliable among all RPs within $C(m)$. Consequently, the corresponding weight $w_{n,l}(t_o)$ can be chosen as the inverse of estimated MED of each RP $n$ as 
    \begin{align} \label{weight}
        w_{n,l}(t_o) = \frac{1}{d_{n,l}(t_o)}.
    \end{align}
    The total set of weights can be defined in a sorted set as  $\bm{w}_l=\{w_{n,l}\!\mid\!{w_{n,l} > w_{n',l}, n<{n'} \,\, \forall n, n' \in C(m)}  \}$, whilst we select the indexes with the first $k$ largest weights to be $\bm{w}_{k,l} = \{w_{1,l},\dots,w_{n,l},\dots,w_{k,l} \,| \, n \in \bm{w}_l \}$. Therefore, the estimated user position $\bm{\hat{x}}(t_o) = [\hat{x}(t_o), \hat{y}(t_o)]$ acquired at time instant $t_o$ can be computed by 
    \begin{align} \label{wknn}
        \bm{\hat{x}}(t_o) = \frac{\sum_{l\in N_{ap}}\sum_{w_{n,l} \in \bm{w}_{k,l}}w_{n,l}\cdot \bm{x}_{n}}{N_{ap}\cdot \sum_{w_{n,l} \in \bm{w}_{k,l}} w_{n,l}},
    \end{align}
    where $\bm{x}_{n}$ is the geometric location of the $n^{th}$ RP. The proposed \WKNN scheme incorporates both cluster information from \RSS and the RSS relationship between the user and RPs. By leveraging the cluster information, RPs having similar RSS values but located farther away are excluded from the top $k^{th}$ weighting elements, which enhances the accuracy of location estimation.
    
\section{Performance Evaluation}\label{section4}
\subsection{Simulation Results}

    \begin{table}[t]
    \begin{center}
    \caption {System Parameters}
        \begin{tabular}{ll}
            \hline
            Parameters of system & Value \\ \hline \hline   
            Number of WiFi APs & 3 \\ 
            Number of RPs & 56 \\ 
            Number of TPs & 89 \\ 
            Indoor topology       & 8$\times$10 $m^{2}$ \\ 
            Size of each grid   & 1.2$\times$1.2 $m^{2}$ \\
            Carrier frequency $f$ & 2.4 GHz \\ 
            Channel bandwidth  & 22 MHz \\ 
            Number of sample per RP  & 10 samples\\ 
            Number of nodes in hidden layers &  $\{ 64, 256, 512, 128, 64 \}$\\
            Learning rate $\eta$ & 0.1 \\
            Threshold $\gamma$ & 1 \\
            Number of nearest neighbor $k$ & 3 \\
	    Pre-training data (sim./exp.) & $\{1120, 4200\}$ samples\\
	   Training data (sim./exp.) &  $\{2240, 8400\}$ samples\\	    
            \hline
        \end{tabular} \label{tab:sim_rss_params}
    \end{center}
    \end{table}

\begin{figure}[t]
        \centering
        \begin{subfigure}[b] {0.15\textwidth}
            \includegraphics[width=\textwidth]{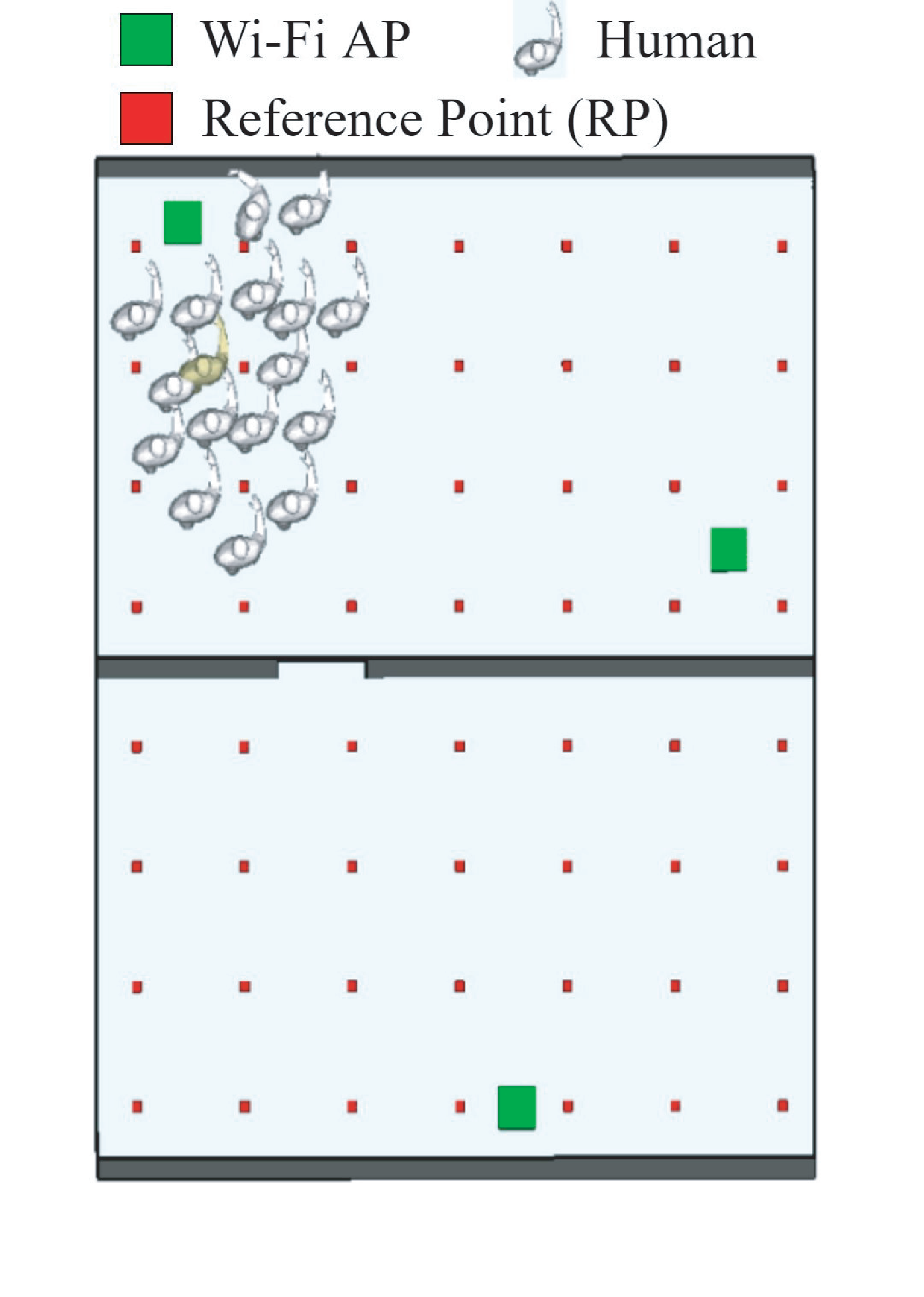}
            \subcaption{}
            \label{fig:deployment_sim}
        \end{subfigure} 
        \begin{subfigure}[b] {0.15\textwidth}
            \includegraphics[width=\textwidth]{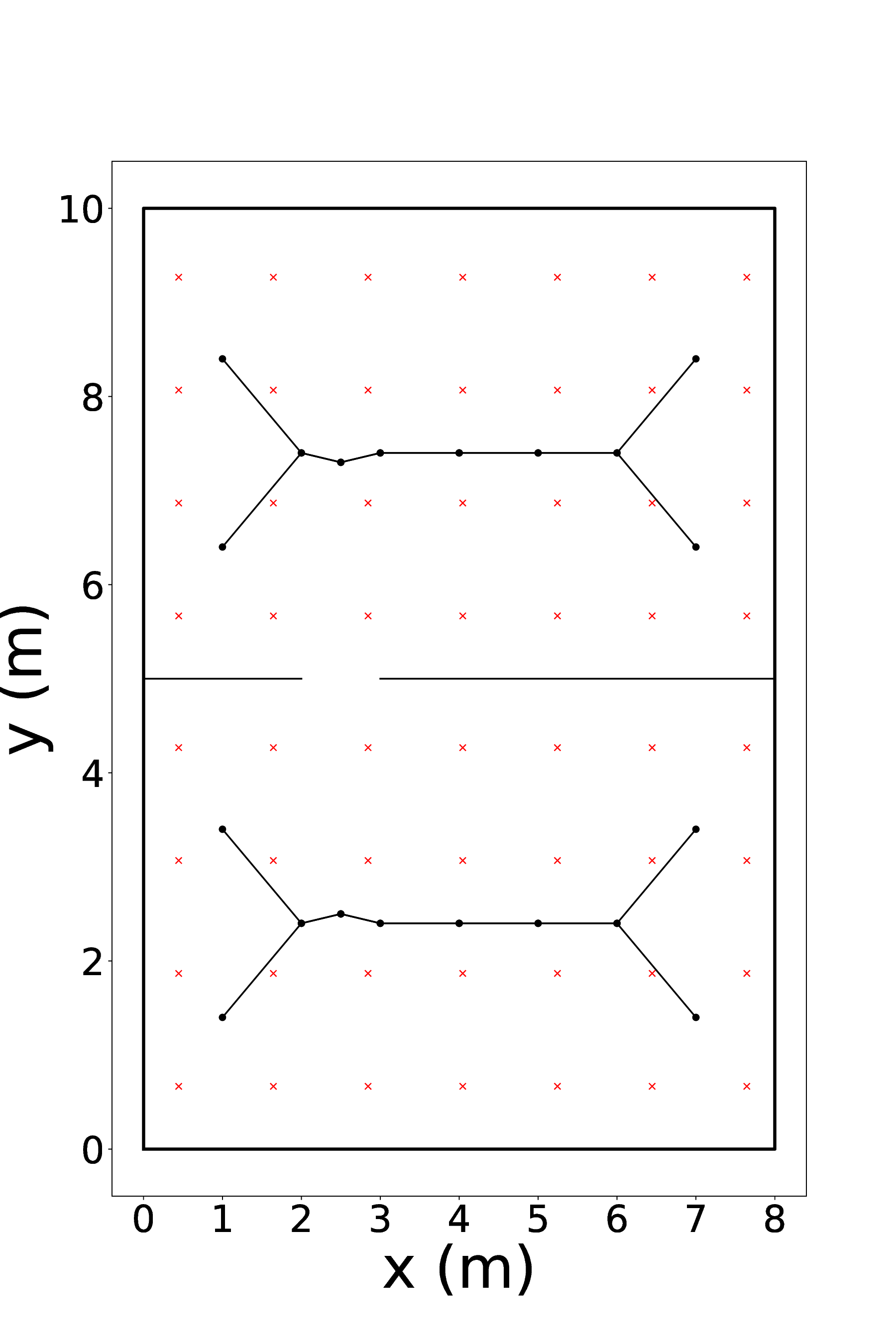}
            \subcaption{}
            \label{fig:skeleton}
        \end{subfigure} 
        \begin{subfigure}[b] {0.15\textwidth}
            \includegraphics[width=\textwidth]{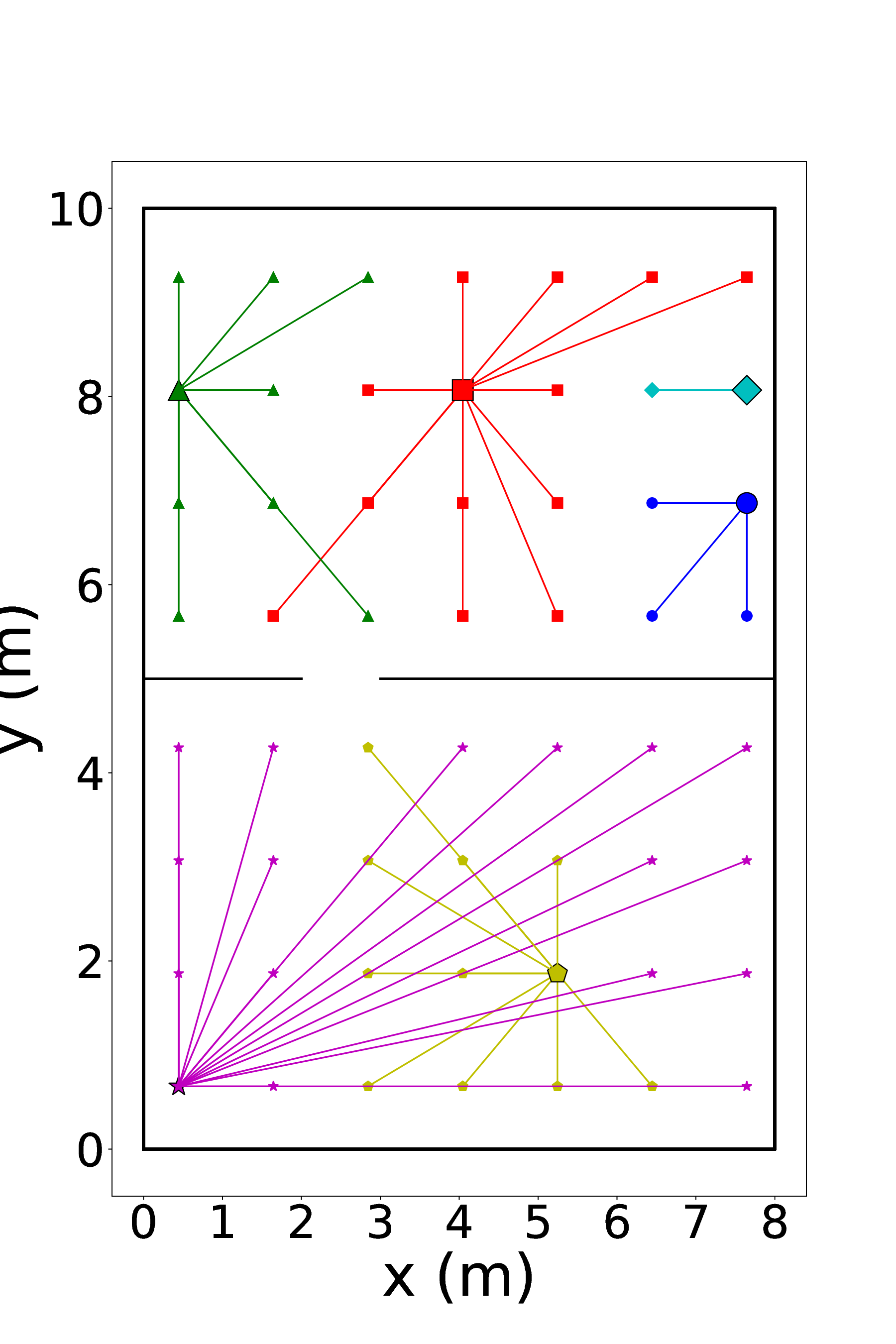}
            \subcaption{}
            \label{fig:cluster_result_RSS}
        \end{subfigure} 
        \newline
        \begin{subfigure}[b] {0.15\textwidth}
            \includegraphics[width=\textwidth]{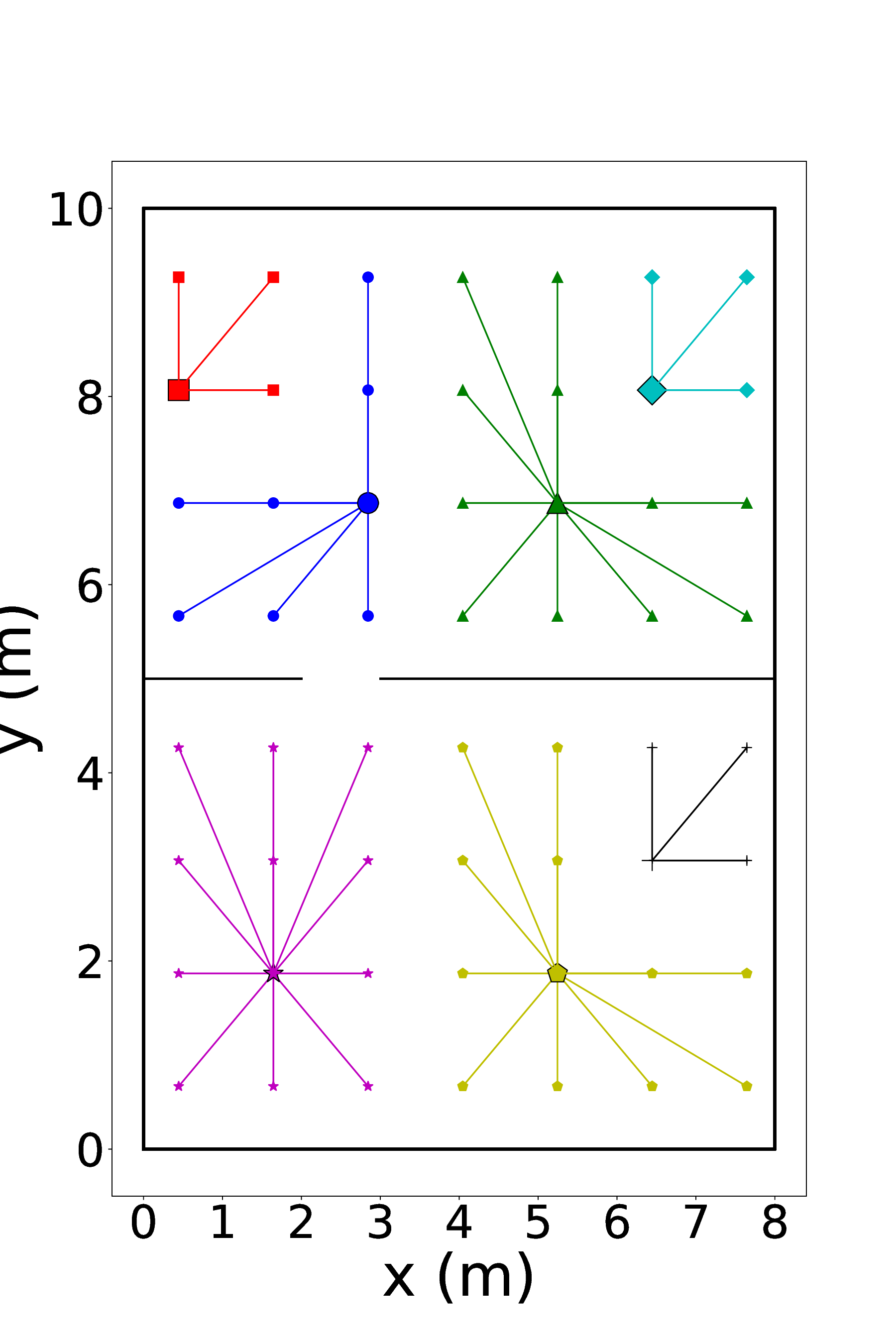}
            \subcaption{}
            \label{fig:cluster_result_SSP}
        \end{subfigure} 
        \begin{subfigure}[b] {0.15\textwidth}
            \includegraphics[width=\textwidth]{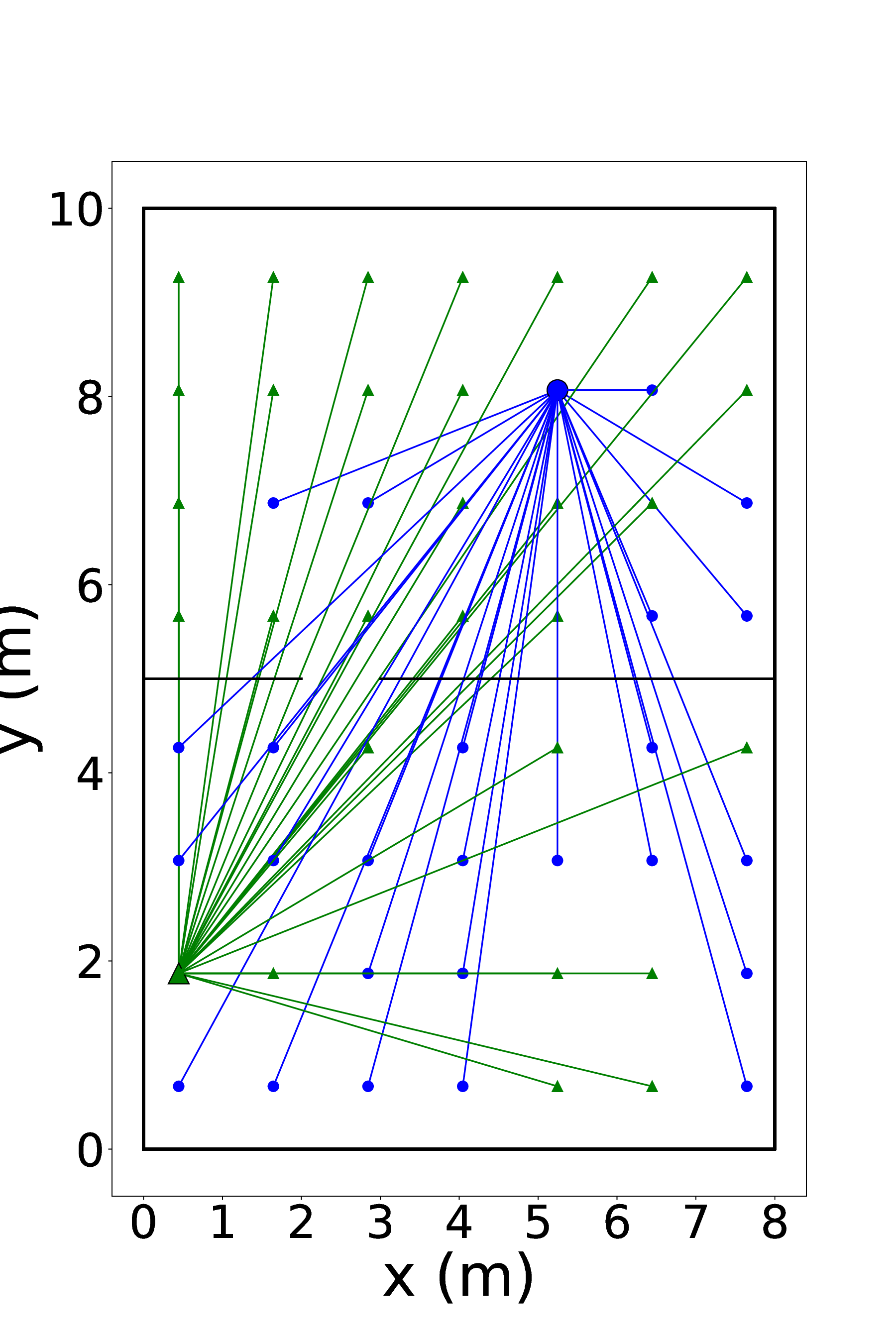}
            \subcaption{}
            \label{fig:cluster_result_timevarying}
        \end{subfigure}
            \begin{subfigure}[b] {0.15\textwidth}
            \includegraphics[width=\textwidth]{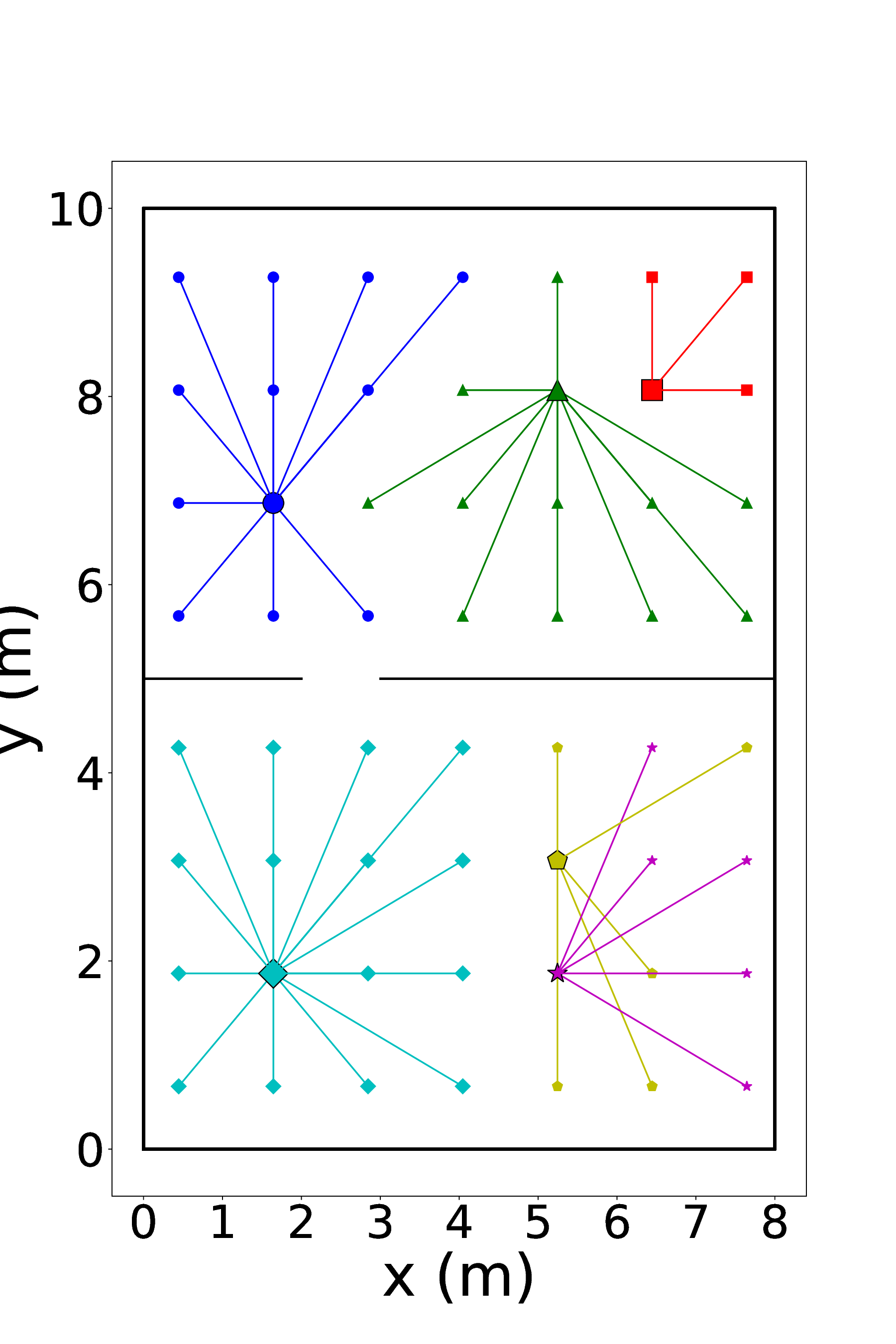}
            \subcaption{}
            \label{fig:sim_clustering}
        \end{subfigure}
        \captionsetup{font={footnotesize}}
        \caption{(a) Network setup with two-room layout for simulations. (b) The generated skeletons from \RSS of proposed \SYS system. The clustering results by considering similarity of (c) the difference of RSS $d_{RSS}(i,j)$ in (\ref{drss});  (d) the shortest path $d_{SSP}(i,j)$ in (\ref{dssp}); (e) the time-varying RSS difference $\delta_{i,j}$ in (\ref{dtimevarying}); and (f) the combined factors $s_{i,j}$ in (\ref{similarity}) of \RSS.}        
        \label{fig:cluster_result}
    \end{figure}

    We firstly evaluate the performances of proposed \SYS system including \RSS, \REG and \WKNN schemes via simulations. We employ Wireless InSite which is a widely-adopted simulation software to emulate ray-tracing based indoor wireless propagations. We consider a two-room scenario with each room size of $8$ $\times$ 10 m$^2$ as shown in \fig\ref{fig:deployment_sim}, where three APs (marked as large green squares) are deployed at the room corners operating at the frequency of $2.4$ GHz. There are $56$ RPs (marked as small red squares) evenly distributed with a inter-RP distance\textsuperscript{\ref{note1}}\footnotetext[1]{In our proposed system, the RPs are placed in a pattern of uniform grids. We empirically find in our several experiment trials that the RP location will not substantially affect the performance of the proposed scheme. The deployment of RPs can potentially enhance the performance, which is proved in some existing works. However, the optmal RP deployment requires a much more complex scheme, which is out of the scope of this paper and can be left as the future work.\label{note1}}of $1.2$ m, whilst $89$ random test points (TPs) are set in both rooms. As depicted in \fig\ref{fig:deployment_sim}, two different cases are considered as follows: (a) both rooms are empty and (b) left part of  top room is crowded with people and bottom room is vacant. We sample $10$ time slots to generate different channel conditions for each RP, and the weights in (\ref{similarity}) is chosen as $\bm{\omega} = [1/3, 1/3,1/3]$. Furthermore, the number of nodes in hidden layers of FCN are designed as $\{ 64, 256, 512, 128, 64 \}$ as shown in \fig\ref{figure:NN}. The learning rate $\eta$ and threshold $\gamma$ in \NN is set as $0.1$ and $1$, respectively. The rectified linear unit (ReLU) is selected as the activation function $f^{(h)}(\cdot)$ in (\ref{deltaRP}). The number of nearest neighbor is set to $k=3$ in \WKNN scheme. The volume of pre-training and training data is 1120 and 2240 samples, which corresponds to 20 and 40 samples per RP, respectively. Table \ref{tab:sim_rss_params} summarizes the parameter setting in simulations.

Table \ref{tab:complexity} elaborates the computational complexity of CODE-LR and CODE-NN compared to the existing method of CSE \cite{CSE}. CSE has the highest complexity order of $\mathcal{O}\left(N^3 \times N_{rp} \times N_{ap} \times \kappa \right)$, where $N$ indicates the data size and $\kappa$ stands for the kernel size. Note that $N^3$ comes from the cross-comparison of input in support vector regression mechanism, whereas $\kappa$ will depend on what kind of kernel is adopted for cross-comparison. The proposed CODE-LR scheme has the lowest computational complexity order of $\mathcal{O} \left(N_{rp} \times N_{ap}\right)$, which is only proportional to the network deployment size $N_{ap}$ as well as measuring points $N_{rp}$. Since the input feature of CODE-LR requires only RSS from the MP, the dimension of input feature becomes $N=1$, which therefore has a lower complexity order than CSE. On the other hand, CODE-NN possesses a moderate computational complexity order of $\mathcal{O} \left(N \times N_{mp} \times \prod_{i=1}^{N_L}{K_{i}^{2}} \times N_{rp}\times N_{ap}\right) $, where additional complexity comes from neural network layers $N_L$ and the corresponding neurons denoted by $K_i$ for the $i$-th layer.

\begin{table}[t]
    \begin{center}
	\caption {Computational Complexity Comparison}
    \begin{tabular}{lr}
	\hline            
	Scheme & Computational Complexity \\
	\hline
	CSE \cite{CSE} &   $\mathcal{O}\left(N^3 \times N_{rp} \times N_{ap} \times \kappa \right)$\\
	CODE-LR & $\mathcal{O} \left(N_{rp} \times N_{ap}\right)$ \\
    CODE-NN & $\mathcal{O} \left(N \times N_{mp} \times \prod_{i=1}^{N_L}{K_{i}^{2}} \times N_{rp}\times N_{ap}\right) $  \\
	\hline     
    \end{tabular} \label{tab:complexity}
\end{center} 
\end{table}

    The SSP-skeleton generated from \RSS algorithm is shown in \fig\ref{fig:skeleton}, where the black lines are skeletons and black dots are vertices. Note that the red crosses represent the RPs to be tested. In \fig\ref{fig:cluster_result_RSS}, the clustering result shows that the RPs are divided into 6 clusters by employing only the difference of RSS $d_{RSS}(i,j)$ in $\eqref{drss}$ among RPs. However, only considering the RSS difference leads to the two clusters in the bottom room overlapping each other with RPs belonging to a cluster located farther away. \fig\ref{fig:cluster_result_SSP} shows the clustering result by only adopting the map information of each RP, i.e., using the shortest path $d_{SSP}(i,j)$ in $\eqref{dssp}$, where all generated clusters will not overlap with each other due to the characteristic of SSP. Nonetheless, the clustering result is unable to reflect wireless signal propagation. \fig\ref{fig:cluster_result_timevarying} demonstrates the clustering results by utilizing the time-varying RSS difference $\delta_{i,j}$ in $\eqref{dtimevarying}$. It reveals that taking time-variation into account will only generate $2$ clusters, causing them to overlap and even across the two-room partition. By implementing the proposed \RSS scheme which considers all factors in  $\eqref{similarity}$, $6$ clusters are automatically generated from all RPs as shown in \fig\ref{fig:sim_clustering} where the cluster heads are chosen as the locations of MPs for the corresponding clusters. The benefits of considering all three factors reveal that each clusters will not largely overlap with the others.

    \begin{figure}[t]
        \centering
        \begin{subfigure}[b] {0.23\textwidth}
        \includegraphics[width= \textwidth]{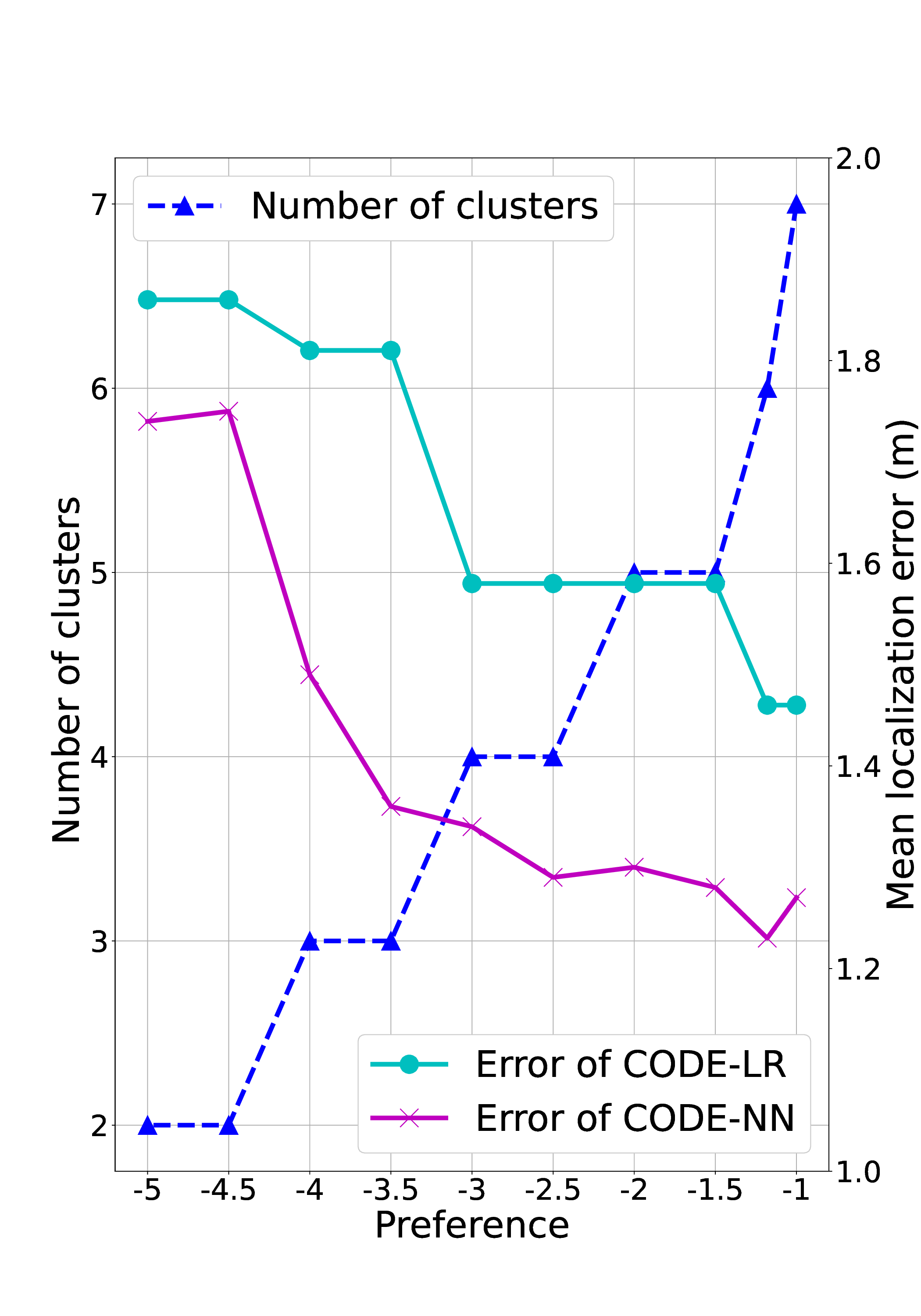}
        \subcaption{}
        \label{fig:ROMAC_comparison_1}
        \end{subfigure}
        \begin{subfigure}[b] {0.23\textwidth}
        \includegraphics[width= \textwidth]{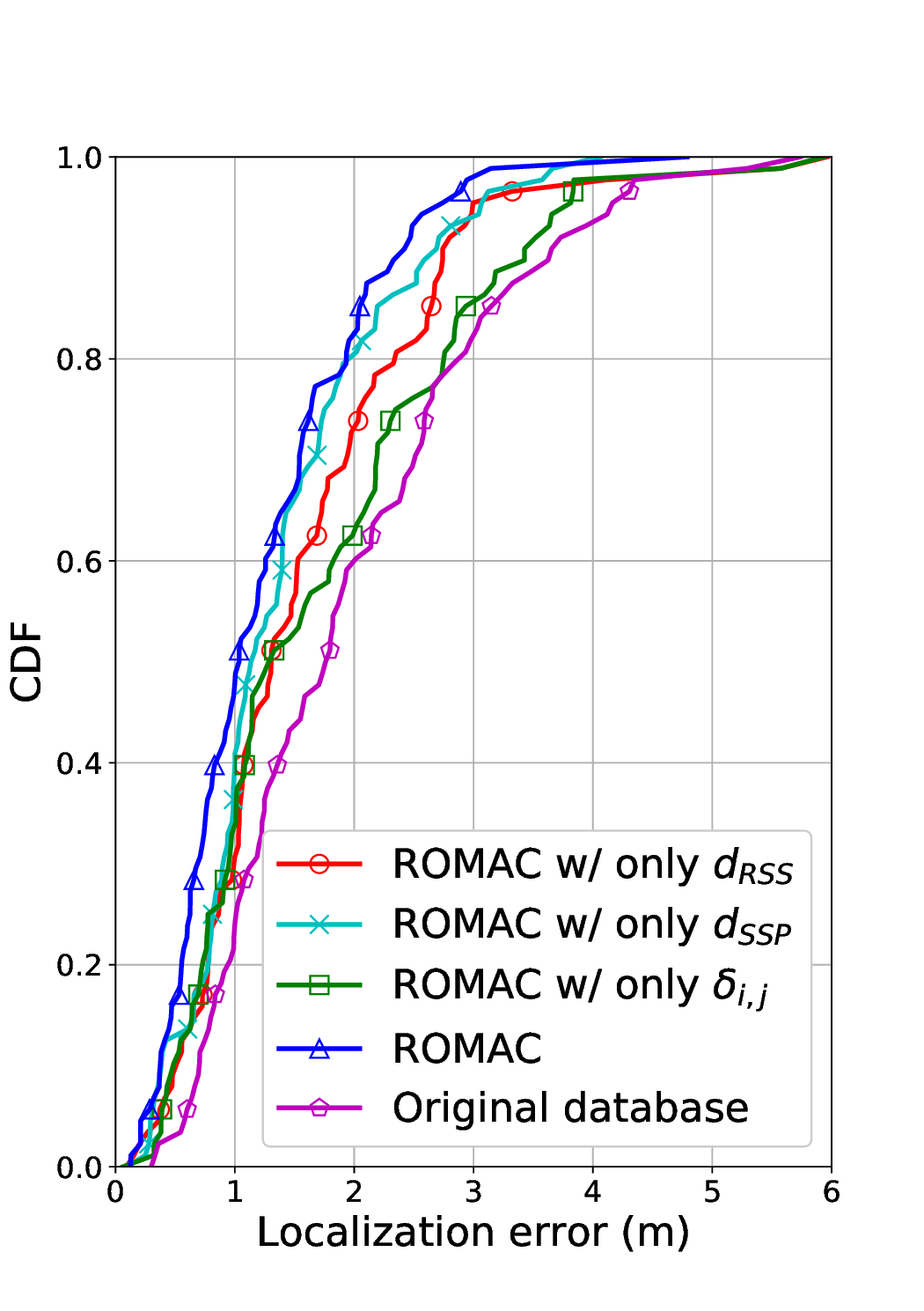}
        \subcaption{}
        \label{fig:ROMAC_comparison_2}
        \end{subfigure}
        \captionsetup{font={footnotesize}}
        \caption{Simulation results of \RSS in SALC system in terms of (a) mean localization errors and corresponding number of clusters using different preference values of \RSS and (b) CDF of localization error by comparing different factors in \RSS.}
        \label{fig:ROMAC_comparison}
    \end{figure}
    
    \fig\ref{fig:ROMAC_comparison} shows the performance evaluation under different parameters of the \RSS scheme. \fig\ref{fig:ROMAC_comparison_1} demonstrates the resulting number of clusters and localization error under different values of self-similarity preference $s_{j,j}$. We evaluate proposed \LR and \NN in terms of localization errors with the corresponding number of clusters from \RSS. It can be observed that a smaller preference value will generate fewer clusters. However, a small number of clusters cannot reconstruct the database efficiently, which leads to a higher localization error in both schemes. The lowest error is reached at the preference value of $s_{j,j}=M_d(\bm{S}_j)=-1.18$ in $\eqref{self-similarity}$ under \NN, which is chosen as our preference value without the limitation of the number of MP. \fig\ref{fig:ROMAC_comparison_2} compares the cumulative distribution function (CDF) of localization error in the crowded environment by using different similarity measures such as difference of RSS amplitude, SSP for map information, and time-variation of RSS, and combined factors using \RSS to choose MPs among the clustered RPs. Note that the databases in these four cases are constructed by \NN, whilst the curve named original database indicates the utilization of fingerprint database in empty environments. It can be seen that the proposed \RSS with adaptive database achieves the lowest localization error, which outperforms the other methods suffering from time-varying signal blockages and reflection. Again, this can be further emphasized with the aid of Fig. \ref{fig:cluster_result}. As illustrated in Figs. \ref{fig:cluster_result_RSS} and \ref{fig:cluster_result_SSP}, using only $d_{SSP}$ for clustering results in separation based on geometric relationships, which neglects the other crucial factors such as $d_{RSS}$ capturing the impact of \emph{path loss caused by indoor environments}. On the other hand, $\delta_{i,j}$ takes into account \emph{dynamic signal strength fluctuations} caused by the presence of people, as shown in Fig. \ref{fig:cluster_result_timevarying}. Disregarding these factors can lead to significant deviations in the estimated positions. To elaborate a little further, the simulations presented in this study provide a simplified scenario for evaluating the performance of the proposed clustering algorithm. It is important to acknowledge that disregarding these critical factors can have even more severe consequences in real-world experiments. Real-world environments impose additional challenges of interference and channel variations that can further affect the performance of positioning systems. Therefore, it becomes compellingly essential to consider these factors when designing and implementing positioning systems in practical scenarios.
   
    \begin{figure}[t]
    \centering
    \begin{subfigure}[b] {0.23\textwidth}
    \includegraphics[width= \textwidth]{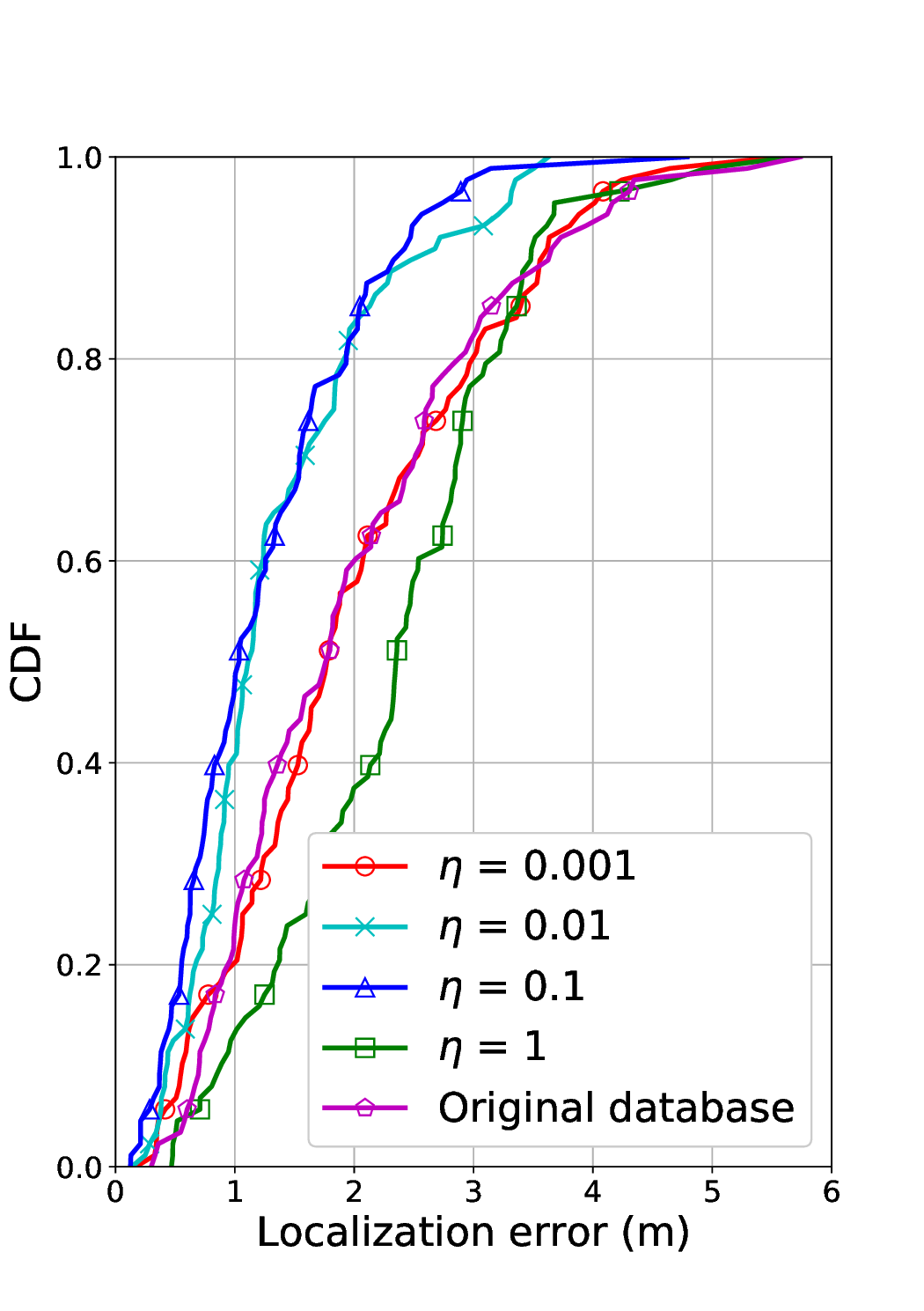}
    \subcaption{}
    \label{fig:eta_compare}
    \end{subfigure}
    \begin{subfigure}[b] {0.23\textwidth}
    \includegraphics[width= \textwidth]{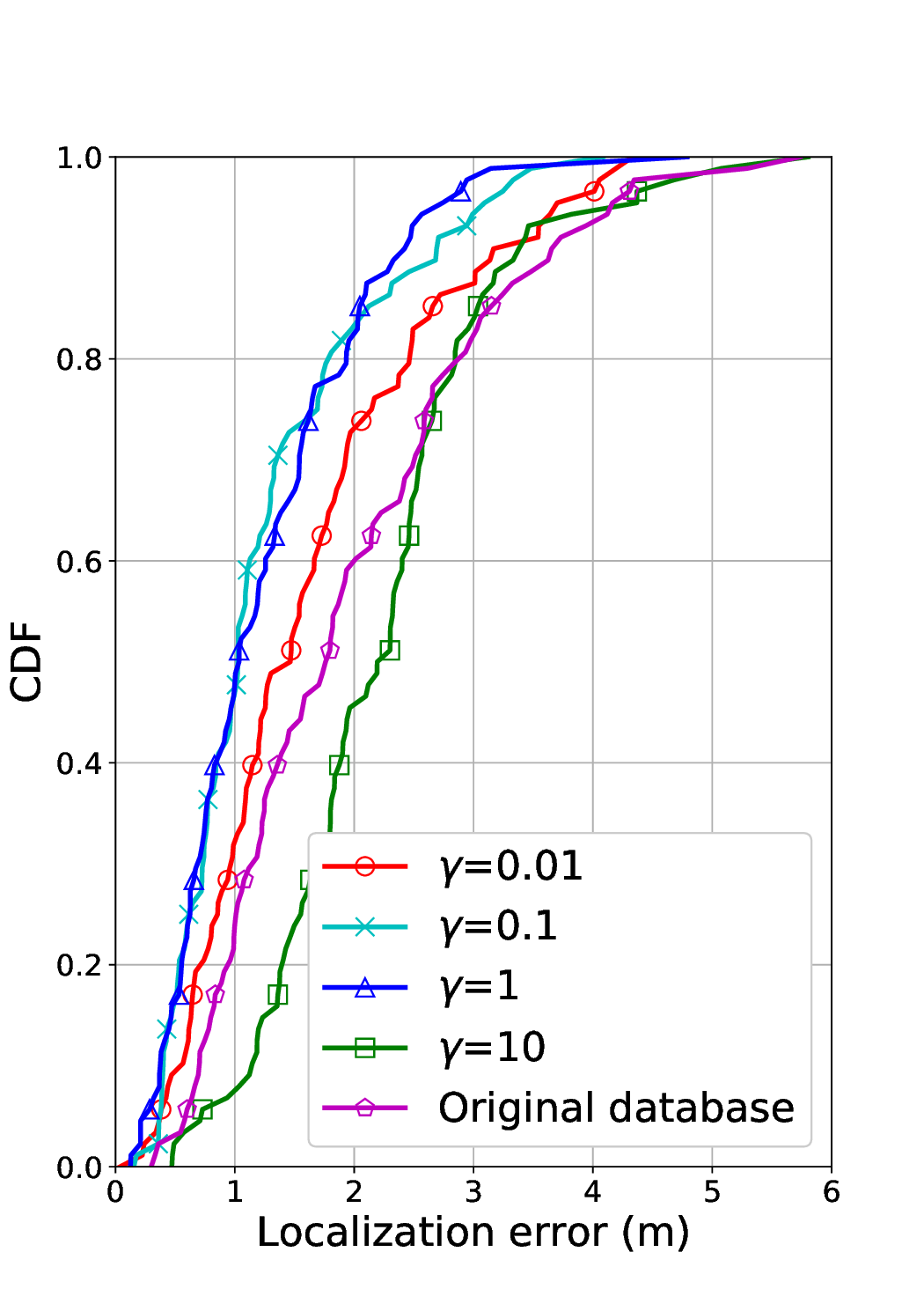}
    \subcaption{}
    \label{fig:gamma_compare}
    \end{subfigure}
    \captionsetup{font={footnotesize}}
    \caption{The CDF of localization error of proposed \NN scheme in SALC system under different (a) learning rates $\eta$ and (b) thresholds $\gamma$.}
    \end{figure}

    \fig\ref{fig:eta_compare} shows the CDF of localization error using different learning rates $\eta$ in \NN to reconstruct the online database. The lines from top to bottom represent $\eta= \{0.001, 0.01, 0.1, 1 \}$ and original fingerprint database, respectively. The result shows that the model has the largest error when $\eta = 1$ since large learning rate potentially diverges the loss function, whilst the database adopting $\eta=0.001$ will be inefficient, since the model over-fitting is induced during data training. Moreover, using the database reconstructed under $\eta=0.1$ can properly converge the loss function and avoid the over-fitting issue, which provides a better performance. Hence, we select $\eta=0.1$ as the learning rate in our \NN scheme in the following simulations. \fig\ref{fig:gamma_compare} shows the CDF of localization error when adopting different thresholds $\gamma$ in the loss function of pre-training in $\eqref{loss1}$ and fine-tuning phases to re-establish the database. The curves include parameters of $\gamma= \{0.01, 0.1, 1, 10 \}$ and original database. The localization error shows that the loss function is unable to filter the outlier when $\gamma$ is set to $\gamma=10$, which causes the reconstructed database unavailable. However, the loss function will treat every output as outlier when we set the threshold as $\gamma=0.01$, which leads to erroneous back propagation of parameters. The CDF of localization error when $\gamma=1$ outperforms the others, which can successfully filter out the outliers. Therefore, we set the threshold as $\gamma=1$ in our purposed \NN scheme. 
    
    \begin{figure}[t]
        \centering
        \begin{subfigure}[b] {0.15\textwidth}
            \includegraphics[width=\textwidth]{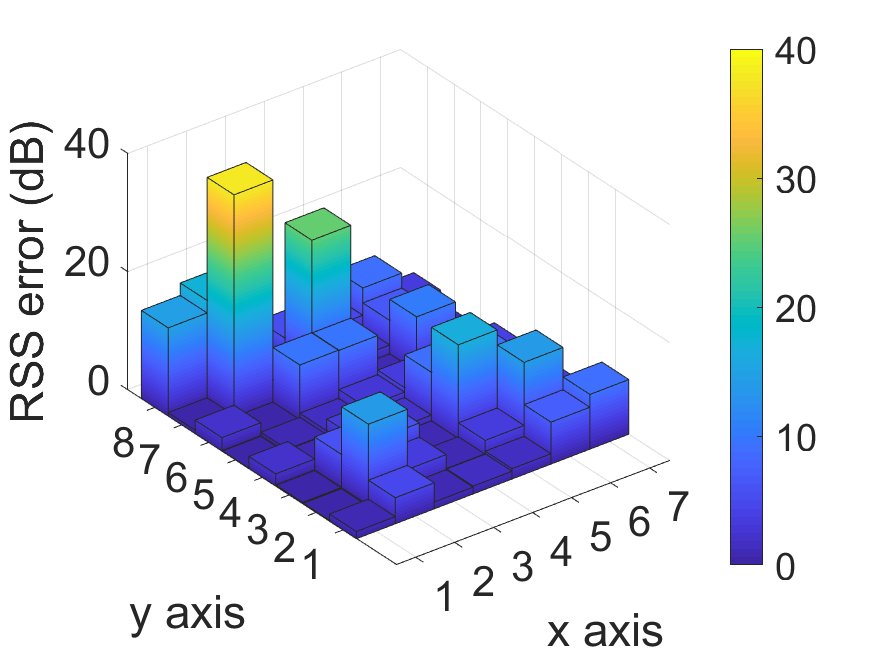}
            \subcaption{}
            \label{fig:RSS_error_original}
        \end{subfigure} 
        \begin{subfigure}[b] {0.15\textwidth}
            \includegraphics[width=\textwidth]{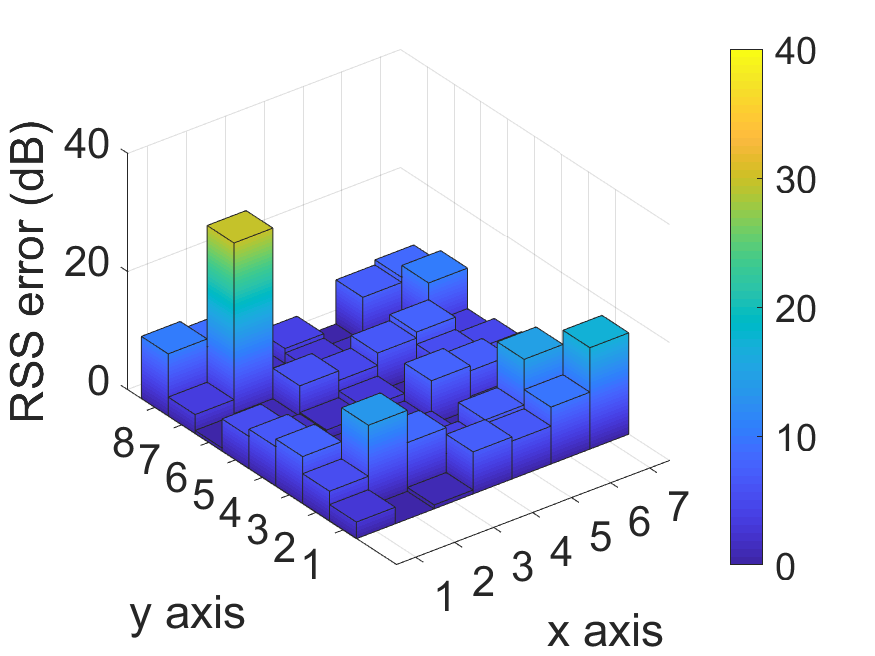}
            \subcaption{}
            \label{fig:RSS_error_reg}
        \end{subfigure}
        \begin{subfigure}[b] {0.15\textwidth}
            \includegraphics[width=\textwidth]{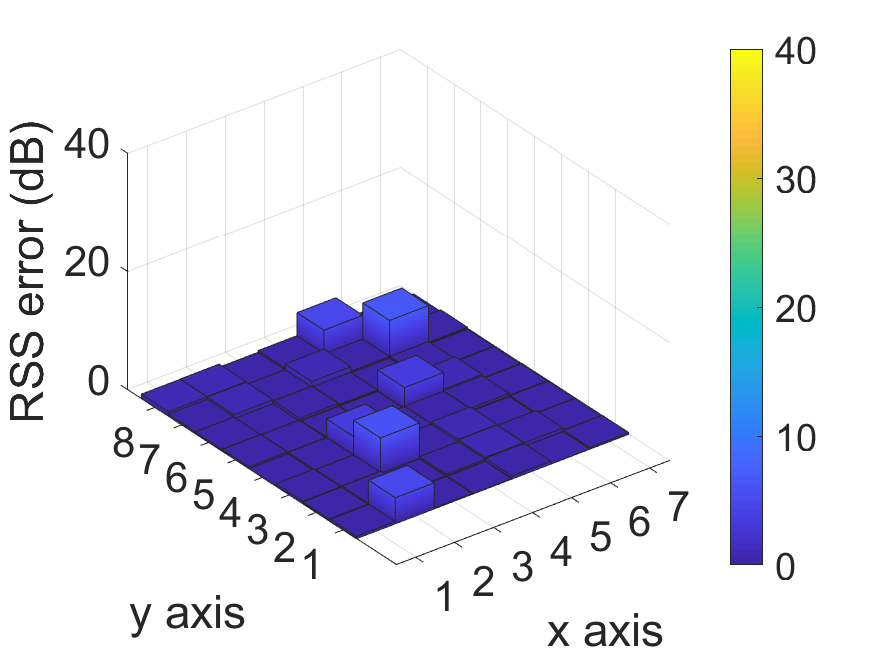}
            \subcaption{}
            \label{fig:RSS_error_nn}
        \end{subfigure}
        \captionsetup{font={footnotesize}}
        \caption{Performance comparison of predicted RSS errors at different RP's locations based on (a) original databased, (b) \LR scheme, and (c)  \NN scheme.}
        \label{fig:RSS_error}
    \end{figure}
    
    \fig\ref{fig:RSS_error} shows the predicted RSS errors at different RPs by adopting original fingerprint database, \LR, and \NN. In all three subplots, the two-dimensional coordinates $[x,y]$ is utilized to represent RP's locations described in \fig\ref{fig:deployment_sim}. The predicted RSS error is calculated as  $\varepsilon_{n,l}(t_o) = |\hat{\alpha}^{RP}_{n,l}(t_o) -\alpha^{RP}_{GT}(t_o)|$, where $\alpha^{RP}_{GT}(t_o)$ represents the ground truth of RSS at $t_o$ and $l=1$ is adopted by using the RSS from the AP located at the upper-left corner of \fig\ref{fig:deployment_sim}. It can be observed that both \NN and \LR methods can reduce RSS error from the original database by reconstructing the adaptive RP database. The largest error can be seen at RP's location $[x,y]=[2,7]$ in \fig\ref{fig:RSS_error_original}, which indicates that the area with crowded people causes higher RSS errors, with the original database collecting RSS under an empty scenario. The result in \fig\ref{fig:RSS_error_reg} shows the regression method in \LR. It can reduce most of RSS errors but has a difficulty to surpress the peak error due to linear operation of regression. \fig\ref{fig:RSS_error_nn} illustrates that the proposed \NN can perfectly reconstruct the radio map with compellingly low RSS errors thanks to its nonlinear mapping in deep neural networks, which outperforms \LR. In addition to propagation decay and cluster information in \LR, \NN considers time-varying effect in different environments, which achieves the lowest localization errors among the other schemes.

    \begin{figure}[t]
    \centering
    \begin{subfigure}[b] {0.47\textwidth}
    \includegraphics[width= \textwidth]{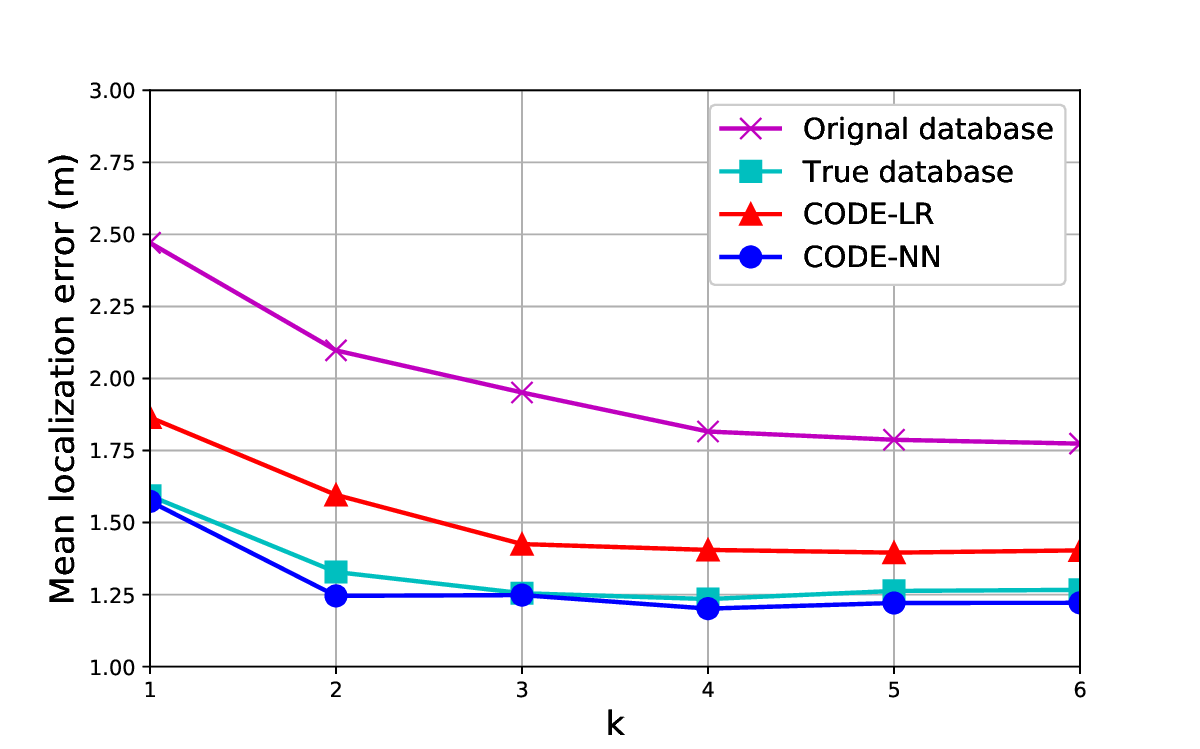}
    \subcaption{}
    \label{fig:k_compare_sim}
    \end{subfigure}
    \begin{subfigure}[b] {0.47\textwidth}
    \includegraphics[width= \textwidth]{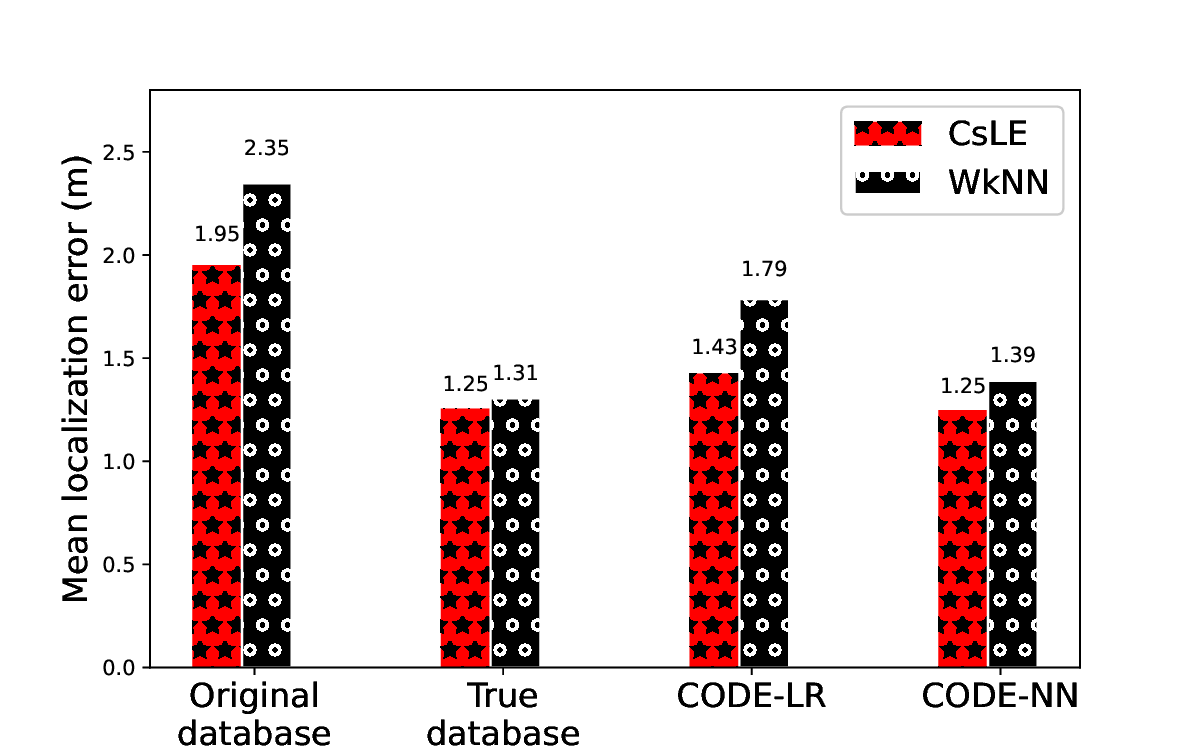}
    \subcaption{}
    \label{fig:sim_result}
    \end{subfigure}
    \captionsetup{font={footnotesize}}
    \caption{Mean localization error of proposed \WKNN scheme (a) under different $k$ values and (b) in comparison with conventional WkNN method.}
    \end{figure}
    
    \fig\ref{fig:k_compare_sim} shows the localization errors by adopting \WKNN scheme in SALC system under different database and $k$ values as indicated in (\ref{wknn}). Note that the original and true databases indicate using the RSS collected from empty and crowded environments, respectively, whilst the crowded one is referred as true database, since it is the realistic case to be dealt with. The result illustrates that using the original database has the largest location error under different $k$ values; while the database reconstructed by \LR and \NN can significantly reduce the error. Additionally, using \NN database is close to adopting the ground truth database, which means we can successfully reconstruct the radio map. It is also shown in the figure that the localization error will decrease with larger $k$ when $k\leq3$, whereas there is no benefit for $k$ larger than $3$ under all four cases. The reason is that a larger $k$ means to take RPs with lower weights into consideration, which are irrelevant to the user's location. Meanwhile, smaller $k$ may cause the chosen RPs to contain insufficient information, which leads to inaccurate predicted location. Accordingly, we select $k=3$ in \WKNN in the following simulations and experiments.
    
    \fig\ref{fig:sim_result} shows the performance comparison between proposed \WKNN and conventional WkNN schemes under $k=3$ with four different types of database. It can be seen that the location errors estimated based on proposed \NN and \LR schemes outperform that from the original database, which suffers from both time-variation and noise. On the other hand, comparably smaller error is generated from true crowded database which is mostly caused by the noise from RSS and RP distances. By taking the time-varying effect into consideration, \NN effectively reconstructs the online database and achieves similar indoor positioning performance compared to that from true database. Meanwhile, \fig\ref{fig:sim_result} also reveals that \WKNN outperforms WkNN with the adoption of those four types of database. For estimating the user's location, WkNN may choose the RPs with similar RSS values even they are located farther away from the user location, whilst our proposed \WKNN will filter those outlier RPs  by adopting the cluster-based feature scaling weight in $\eqref{weight}$. 
    
    \begin{figure}[t]
        \centering
        \includegraphics[width= 3.3in]{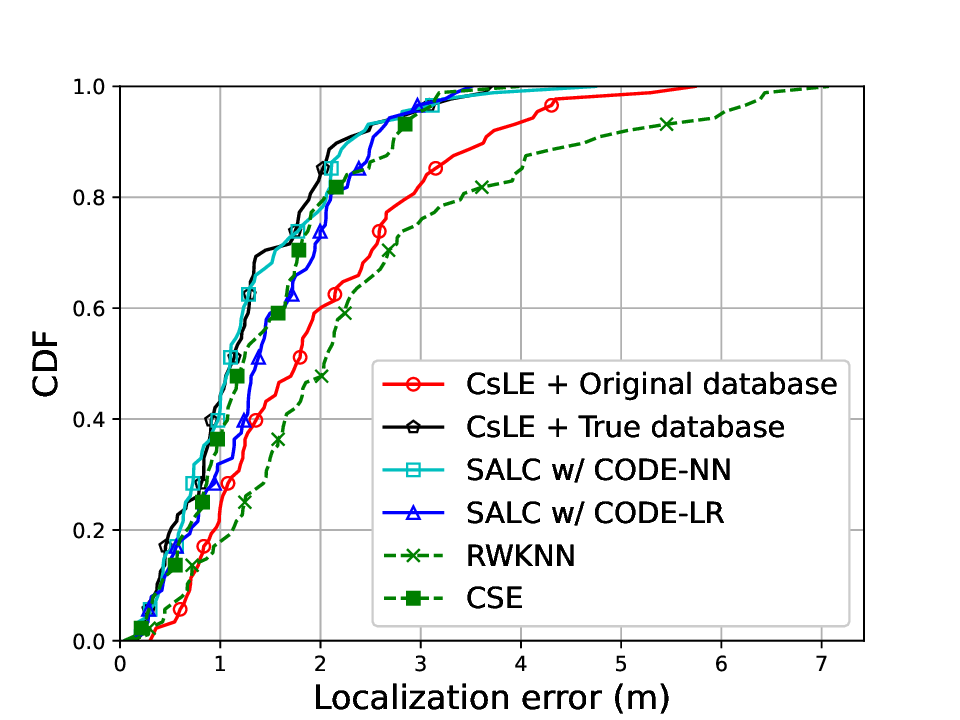}
        \captionsetup{font={footnotesize}}
        \caption{Performance comparison for CDF of localization error under different schemes.}
        \label{fig:sim_result_cdf}
    \end{figure}

 \fig\ref{fig:sim_result_cdf} illustrates the performance comparison on the CDF of localization errors among CsLE with original and true database, SALC with CODE-NN and CODE-LR, and existing schemes from \cite{RWKNN} and \cite{CSE}. Note that RWKNN adopts the original database and CSE reconstructs the database via conventional WkNN. We can observe that the RWKNN method performs the worst performance with localization error of around 2 m under CDF of $0.5$ since it did not consider time-varying effects. Although the original database still encounters time-varying problem, the CsLE scheme can effectively mitigate the problems of both signal fluctuation and selecting farther RPs as neighbor nodes. Additionally, it reveals that the localization error decreases as the fingerprinting database is updated. Furthermore, our proposed SALC scheme achieves the lowest localization error of approximately 1 m under CDF of $0.5$ by dynamically generating the fingerprinting database in time-varying environments.

\subsection{Experiment Results}
   
\begin{figure}[t]
    \centering
    \begin{subfigure}[b] {0.30\textwidth}
    \includegraphics[width=\textwidth]{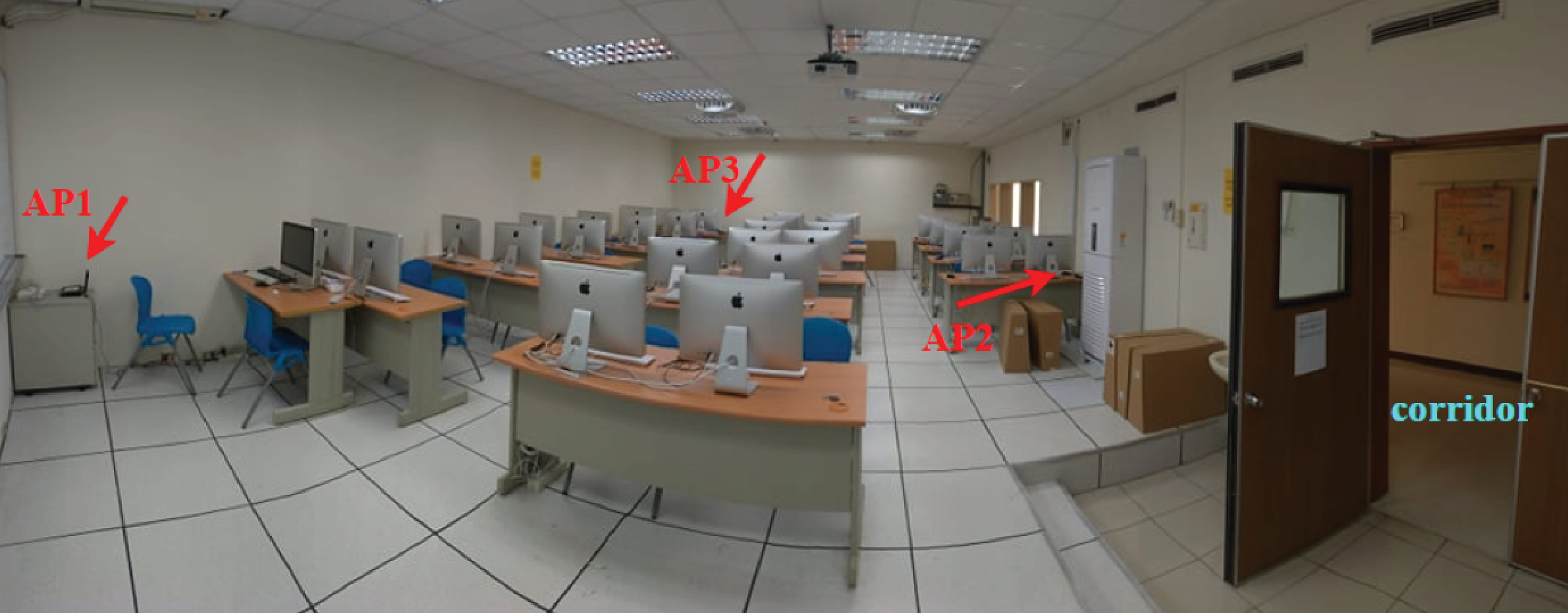}
    \caption{}
    \label{fig:exp_free}
    \end{subfigure} 
    \centering
    \begin{subfigure}[b] {0.16\textwidth}
    \includegraphics[width=\textwidth]{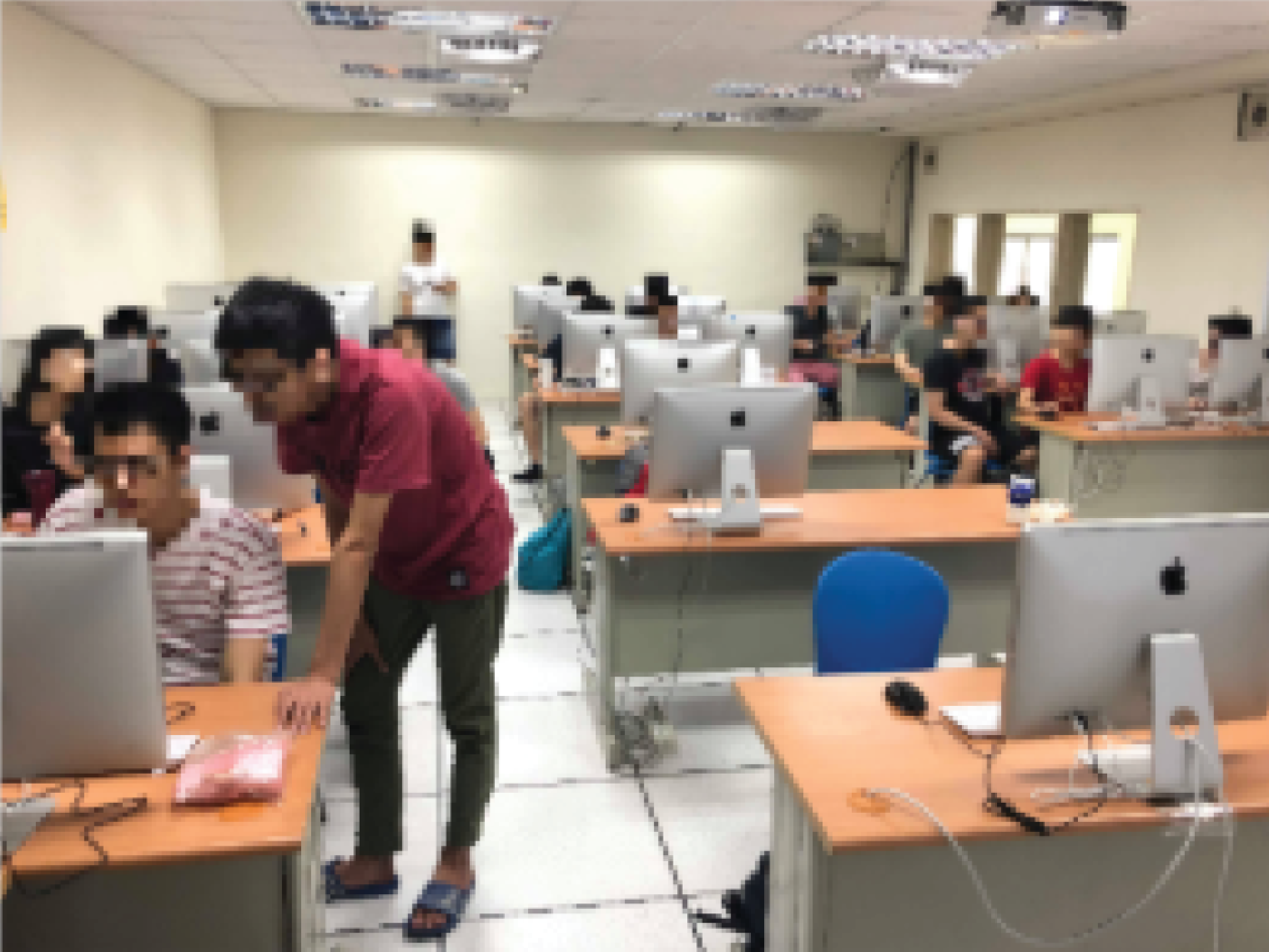}
    \caption{}
    \label{fig:exp_crowd}
    \end{subfigure}
    \captionsetup{font={footnotesize}}
    \caption{Testing field for experiments including both classroom and corridor at (a) break time and (b) in-class time.}
    \label{fig:picture_exp}
    \end{figure}

    Experiments have been conducted to verify the effectiveness of \SYS system in realistic environments. \fig\ref{fig:picture_exp} shows the testing field of experiments including both the classroom and the corridor. We consider break and in-class time representing empty and crowded cases, respectively. The size of the experimental scene \textsuperscript{\ref{note2}}\footnotetext[2]{If the new scene has a different layout, it will be necessary to redeploy the RPs to new locations. This process involves re-clustering the RPs and re-selecting the MPs. Consequently, data from different scenarios in the new scene will need to be collected to adapt ROMAC and CODE algorithms. As a future extension, transfer learning and domain adaptation approaches can be leveraged to address the challenges of retraining the system for different layouts.\label{note2}} is $9.65 \times 10.65$ m$^2$, where $42$ RPs are distributed with inter-RP distance of $1.2$ m for database collection, and $26$ TPs are determined to evaluate the positioning accuracy. We use the mobile device of ASUS Zenfone to collect RSS on each RP from the $3$ APs, which are ASUS RT-AC66U operating at $2.4$ GHz. We use the same model of mobile devices serving as MPs during the online phase, which means that there is no additional functional requirement and overhead for deploying MPs in our experiments. The server employs these RSS values to generate an updated radio map and to estimate the user location, which is therefore transmitted back to the mobile device. Since most of training and computing are conducted at server side, the mobile device collecting data and received positioning results has negligible computation during the process. We collect $100$ samples on each RP in both break and in-class time, which takes around $1$ hour in each case. Note that the crowded true database is not feasible to be collected in practical scenarios, and we establish it mainly to serve as the ground truth for performance comparison. The amount of pre-training and of training data is 4200 and 8400, respectively.
The other system parameters are chosen to be the same as those in simulations.

\begin{figure} [t]
	\begin{subfigure} {.485\linewidth}
		\includegraphics[width=\textwidth]{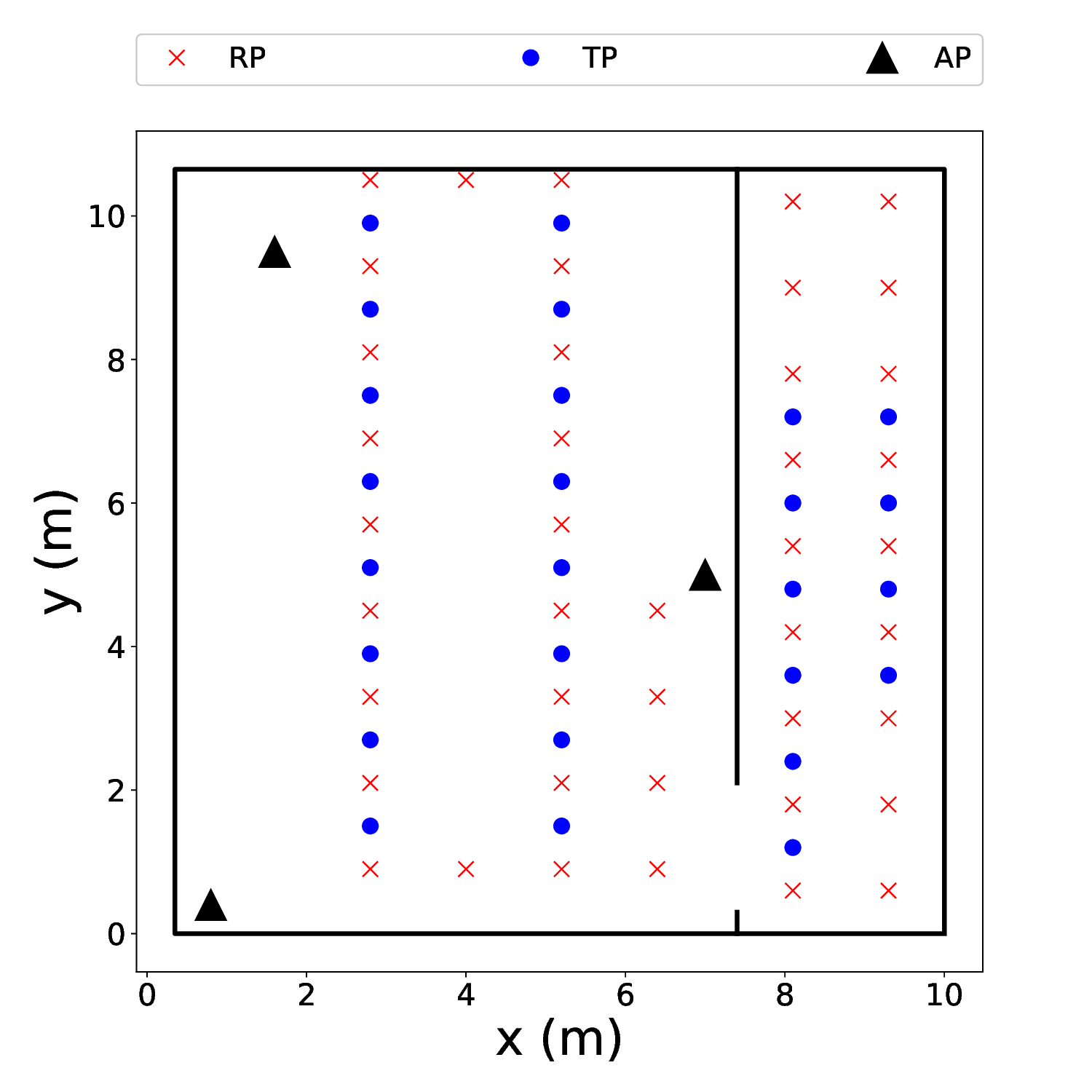}
		\caption{} \label{fig:deployment_exp}
	\end{subfigure}
	\hspace{1mm}
	\begin{subfigure} {.485\linewidth}
		\includegraphics[width=\textwidth]{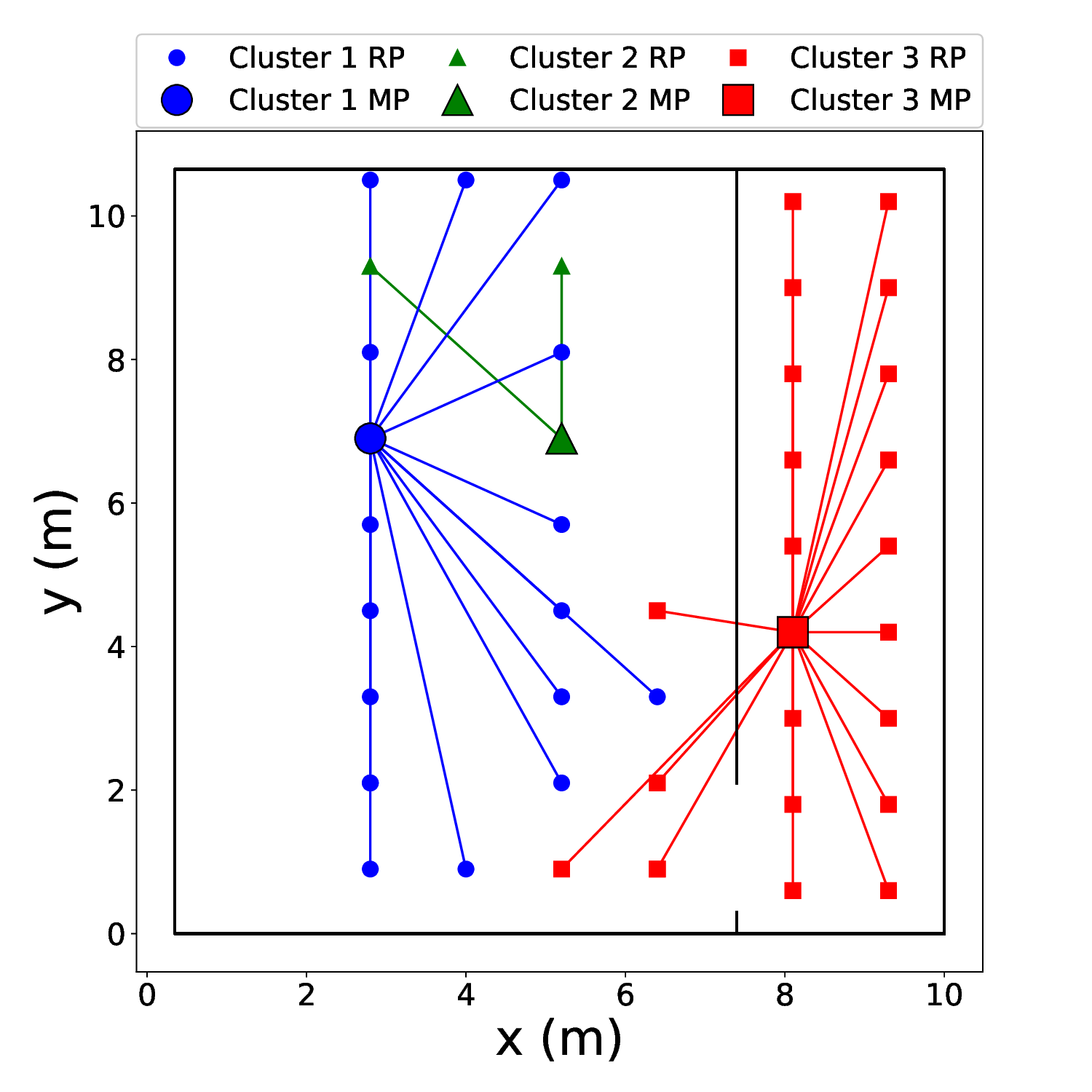}
		\caption{} \label{fig:clustering_exp}
	\end{subfigure}
	\captionsetup{font={footnotesize}}
	\caption{(a) The experimental scene and deployments of APs, RPs, and TPs, and (b) the resulting $3$ clusters and corresponding MPs by adopting proposed \RSS algorithm. }
\end{figure}

    \fig\ref{fig:deployment_exp} illustrates the layout and deployments of APs (black triangles), RPs (red crosses), and TPs (blue points), whilst \fig\ref{fig:clustering_exp} shows the result by adopting proposed \RSS algorithm. With the consideration of hardware limitation in a practical scenario, we adjust the preference value $s_{j,j}=-5$ in \eqref{self-similarity} such that the resulting number of clusters will be limited to $3$. It can be seen from \fig\ref{fig:clustering_exp} that all RPs on the corridor are in the same cluster since the SSP information in (\ref{dssp}) is taken into account in \RSS. The RPs in the testing classroom are divided into two clusters due to human blockage effects, which will reflect the time-varying RSS as considered in \RSS.
   
    \begin{figure} [t]
	\begin{subfigure} {0.48\linewidth}
		\centering
		$\vcenter{\hbox{\includegraphics[width=\textwidth]{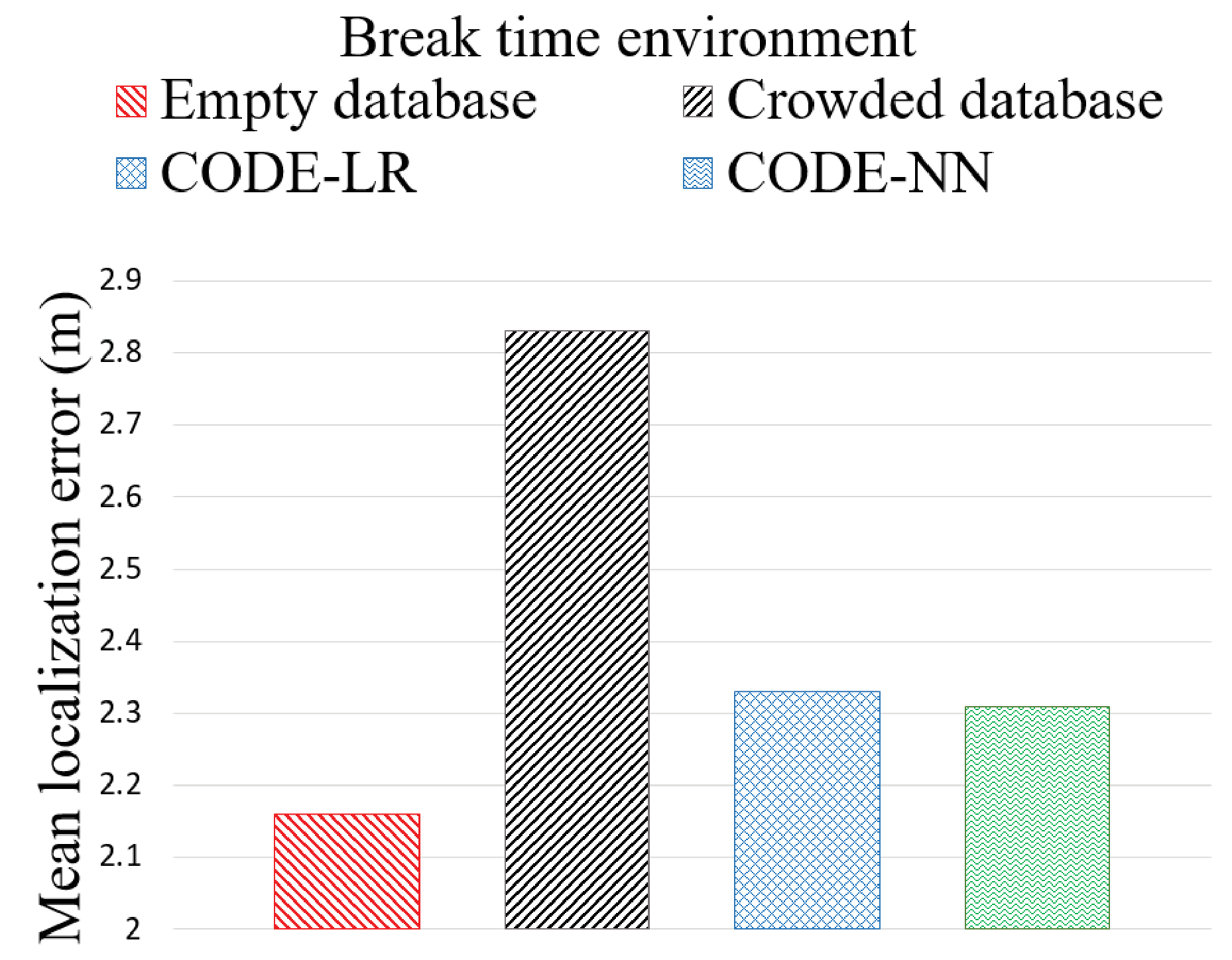}}}$
		\caption{}\label{fig:exp_break}
	\end{subfigure}
	\begin{subfigure} {0.48\linewidth}
		\centering
		$\vcenter{\hbox{\includegraphics[width=\textwidth]{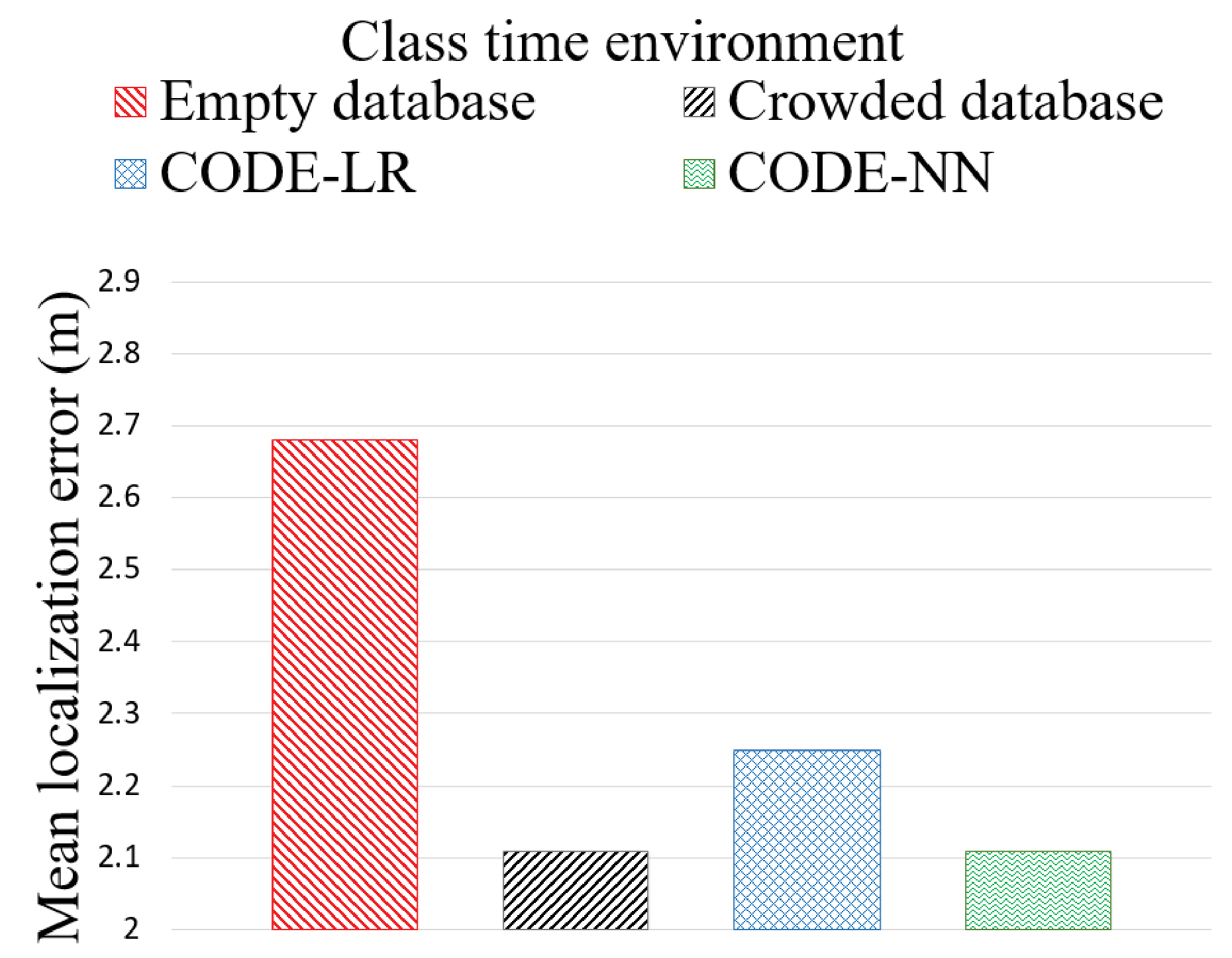}}}$
		\caption{}\label{fig:exp_class}
	\end{subfigure}
	\captionsetup{font={footnotesize}}
	\caption{Performance comparison of proposed SALC system under different types of database during (a) break time and (b) in-class time.\label{fig:exp_result}}
    \end{figure}

    Figs. \ref{fig:exp_break} and \ref{fig:exp_class} show the positioning results of proposed SALC system in real-time environment. The cases of using empty and crowded databases indicate that the RSS data are collected in break time and in-class time, respectively. \fig\ref{fig:exp_break} shows the results of noise-free environment, where the mean error of using crowded database is the largest of $2.83$ m among the other databases since the RSS from the in-class time database suffers from human blocking effects. Note that the empty database with localization error of $2.16$ m is treated as the ground truth database in break time. It can be observed that the errors from proposed \LR and \NN are respectively reduced to $2.31$ and $2.33$ m compared to that acquired from crowded database during in-class time. Furthermore, \fig\ref{fig:exp_class} illustrates the experimental results under noisy environments, where the mean localization error of using the empty database is the largest of $2.68$ m since the database was not collected with signal blockage features. Note that the crowded database is treated as the ground truth one in this case resulting in $2.11$ m of positioning error. It can be seen that the proposed \LR method achieves the error of  $2.25$ m, whilst \NN results in the same smallest $2.11$ m error as that from ground truth crowded database.

    \begin{figure}[t] 
        \centering
        \includegraphics[width=3.3in]{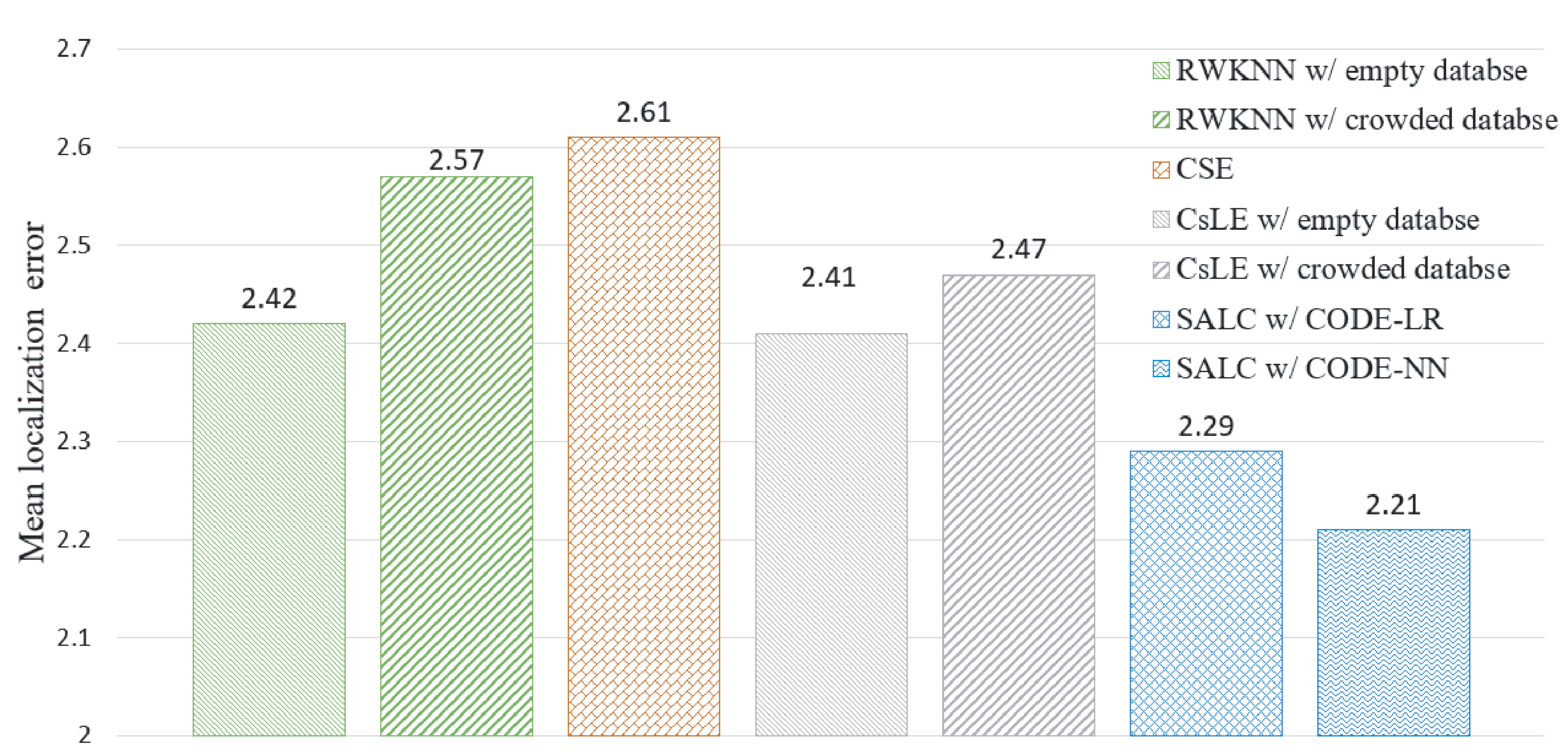}
        \captionsetup{font={footnotesize}}
        \caption{Performance comparison of proposed SALC system with existing  schemes.\label{fig:exp_all}}
    \end{figure}

    \fig\ref{fig:exp_all} shows the overall performance comparison of proposed SALC system with existing RWKNN \cite{RWKNN} and CSE \cite{CSE}  methods, where the results of mean localization error are averaged test data from both noise-free and noisy environments. The bars from left to right are RWKNN utilizing empty and crowded databases, traditional WkNN with the database constructed by CSE,  proposed \WKNN with empty and crowded databases, and \SYS adopting \LR and \NN to reconstruct the databases. It can be seen that \WKNN outperforms conventional RWKNN under both empty and crowded databases. Note that the localization errors are higher in crowded than empty database for both schemes. The main reason is that RSS collected for crowded data can possess high signal variations due to human movements in realistic scenarios, which incurs infeasible establishment of crowded database. Furthermore, the proposed \LR and \NN methods can improve localization performance by adaptively adjusting and reconstructing databases for different environments. Notice that the CSE method only constructs the database by current information with the worst performance due to the complicate multipath effects. Meanwhile, \NN performs better than \LR since it considers the nonlinear features of time-varying effects between noise-free and noisy environments.

\section{Conclusion}\label{section5}
    In this paper, we have designed an \SYS system for indoor positioning including \RSS, \REG, and \WKNN sub-algorithms. \RSS is designed to simultaneously solve the RP clustering and MP deployment in order to monitor the time-varying problem and reconstruct the adaptive fingerprinting database. \REG establishes adaptive online database by employing linear regression and neural network techniques based on the cluster information from \RSS scheme. At last, \WKNN predicts user position by matching the user's real-time RSS with the adaptive database based on cluster information and predicted signal variation. Although we can reconstruct the radio map precisely, there still exist real-time uncertainties in experiments, such as multipath, noise, and interference from in-class and break time, which limits the performance from theory to implementation. In the future work, we will consider complicated environmental scenarios for further performance enhancement. Nevertheless, the merits of proposed \SYS system can still be observed from both simulations and experiments by improving the performance of fingerprinting-based localization under practical time-varying environments.

\bibliographystyle{IEEEtran}
\bibliography{ref}
\vspace{-1.2cm}

\end{document}